\PassOptionsToPackage{table}{xcolor}
\documentclass{article} % For LaTeX2e
\usepackage{iclr2027_conference,times}
\iclrfinalcopy
\usepackage{amsmath,amsfonts,bm}

\def\eqref#1{equation~\ref{#1}}
\def\1{\bm{1}}

\DeclareMathAlphabet{\mathsfit}{\encodingdefault}{\sfdefault}{m}{sl}
\SetMathAlphabet{\mathsfit}{bold}{\encodingdefault}{\sfdefault}{bx}{n}

\usepackage{hyperref}
\usepackage{url}
\usepackage{graphicx} % DO NOT CHANGE THIS
\usepackage{natbib}  % DO NOT CHANGE THIS AND DO NOT ADD ANY OPTIONS TO IT
\usepackage{caption} % DO NOT CHANGE THIS AND DO NOT ADD ANY OPTIONS TO IT
\usepackage{algorithm}
\usepackage{algorithmic}

\usepackage{newfloat}
\usepackage{listings}
\DeclareCaptionStyle{ruled}{labelfont=normalfont,labelsep=colon,strut=off} % DO NOT CHANGE THIS
\floatstyle{ruled}
\newfloat{listing}{tb}{lst}{}
\floatname{listing}{Listing}

\usepackage{booktabs}

\usepackage{amsmath}
\usepackage{amssymb}
\usepackage{booktabs}
\usepackage{multirow}
\usepackage{algorithm}
\usepackage{algorithmic}
\usepackage{tikz}
\usepackage{placeins}
\newcommand{\tb}[1]{\textbf{#1}}
\usetikzlibrary{arrows.meta,positioning,shapes.geometric}

\title{Flow-Support-Weighted Proposals for\\ Test-Time Verified VLA Policies}

\author{Chen Zhao\textsuperscript{\normalfont{1,2,3$\ast$}},
  Zipeng Ye\textsuperscript{\normalfont{3$\ast$}},
  Haitao Shen\textsuperscript{\normalfont{1,2}},
  Guanlin Li\textsuperscript{\normalfont{1,2}},
  Yuchen Xie\textsuperscript{\normalfont{3}},
  Jing Zhang\textsuperscript{\normalfont{1,2$\dagger$}} \\
  \textsuperscript{1} Renmin University of China, Beijing 100872, China\\
  \textsuperscript{2} Key Laboratory of Data Engineering and KnowledgeEngineering, MOE, China\\
  \textsuperscript{3} Meituan, Beijing 100102, China\\
  \texttt{<2021201645, zhang-jing>@ruc.edu.cn}\\
}

\begin{document}

\maketitle
\fancyhead{}
\begingroup\def\thefootnote{$\ast$}\footnotetext{Equal contribution.}\endgroup
\begingroup\def\thefootnote{$\dagger$}\footnotetext{Corresponding author.}\endgroup
\begin{abstract}
Test-time sampling promises to turn one action prediction into a set of distinct executable candidates that a verifier can choose among. Downstream fine-tuning works against that promise. A vision-language-action model trained to solve one task distribution concentrates its flow-matching decoder around the dominant demonstrated action, so repeated samples differ only marginally and the verifier receives many copies of one decision. We call this \emph{proposal contraction} and identify it as a key constraint on test-time scaling: repeated samples can remain highly redundant as the candidate budget grows. Action Dynamic Sampling (ADS) alleviates the constraint on the proposal side, during training. At each training step, ADS integrates $K$ additional trajectories from the current policy under fresh base noise, scores each by the flow support along its path with a Hutchinson trace estimator, and corrects the policy's local action distribution toward this weighted set through a detached finite-set KL term. Training cost rises; deployment cost is unchanged. Across simulation and four physical-robot tasks, ADS raises proposal spread without the success penalty that Gaussian perturbation incurs, lifts oracle best-of-$8$ success on RoboCasa from 57.6\% to 63.0\%, improves fixed verifiers by 2.4 to 4.9 points in simulation, and raises TokenPB by up to 13.0 points on real-world manipulation.
\end{abstract}

\section{Introduction}

% modified by lyh
% Test-time scaling for robot policies has a simple promise: sample several actions, verify them, and execute the best one \citep{jang2025verifier, ichter2018learning, elbanhawi2014sampling}. This promise is attractive for vision-language-action (VLA) models because downstream manipulation tasks are brittle; a single slightly wrong grasp pose, insertion direction, or waypoint can decide the whole episode \citep{zhao2026action, liu2025hybridvla}. Yet in practice, more samples often buy surprisingly little. The problem is not only that the verifier may be imperfect. The deeper problem is that the policy frequently gives the verifier a nearly degenerate candidate set.

Test-time scaling for robot policies rests on a simple premise: sample several actions, verify them, and execute the best one \citep{jang2025verifier, ichter2018learning, elbanhawi2014sampling}. This is particularly attractive for vision-language-action (VLA) models because downstream manipulation tasks are brittle: a slightly incorrect grasp pose, insertion direction, or waypoint can determine the outcome of an entire episode \citep{zhao2026action, liu2025hybridvla}. Yet in practice, drawing more samples often yields surprisingly little gain. The limitation is not only an imperfect verifier. More fundamentally, the policy often supplies the verifier with a highly redundant, nearly degenerate candidate set.
\begin{figure}
  \centering
  \includegraphics[width=\textwidth]{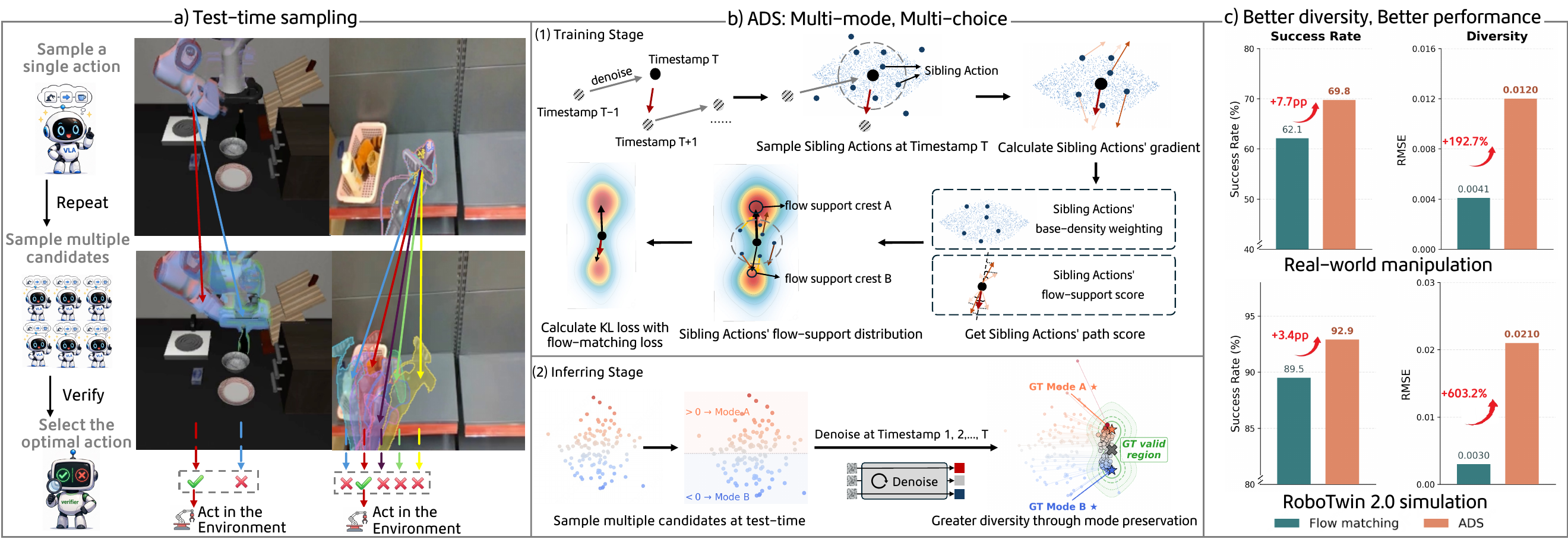}
  \caption{\textbf{ADS produces broader, more useful proposal sets.}
  Standard flow-matching training maps different noise draws onto one dominant action mode, leaving the verifier with redundant candidates.
  ADS corrects the policy against a flow-support-weighted set of its own samples; the resulting candidates cover more valid modes and get better performance, with the verifier unchanged.}
  \label{fig:motivation}
\end{figure}

% This degeneracy is a direct consequence of how VLA models are fine-tuned. Downstream success rewards extreme convergence onto the dominant demonstrated action, which is exactly what single-sample execution wants and exactly what test-time sampling cannot survive \citep{kwok2025robomonkey, dai2025rover}. Once the flow-matching decoder has contracted onto one narrow mode, drawing $M$ action chunks yields one choice with $M$ small perturbations. No verifier can select an alternative grasp, waypoint, or recovery behavior that the proposal model never produced \citep{wang2026sampling}. Injecting Gaussian noise at test time buys geometric spread at the cost of the action manifold: the perturbed chunks are farther apart and less often executable, and our measurements confirm the trade is a losing one.

This degeneracy arises naturally from downstream fine-tuning of flow-matching VLA policies. 
Repeated supervision drives the policy to concentrate around the dominant demonstrated action, which benefits single-sample execution but undermines the diversity that test-time sampling relies on \citep{kwok2025robomonkey, dai2025rover}. 
Once the flow-matching decoder has contracted onto a narrow mode, drawing \(M\) action chunks yields effectively one decision represented by \(M\) minor variations. 
A verifier cannot recover an alternative grasp, waypoint, or recovery behavior that the proposal model never produces \citep{wang2026sampling}. 
Injecting Gaussian noise at test time buys geometric spread at the cost of the action manifold: the perturbed chunks are farther apart and less often executable, and our measurements confirm the trade is a losing one.

% The fix therefore belongs in training, while the policy still has variation to preserve \citep{faria2025sample, bai2025evolve, zhang2026ttt}. The obstacle is that flow-matching decoders expose no normalized probability over candidate actions, so entropy and distribution-matching regularizers have nothing to attach to \citep{zhu2026can}. But a flow model does expose something else. Its velocity field has a divergence at every point along an integration path, and integrating that divergence measures how strongly the learned flow contracts toward the action at the end of the path. We call this quantity the \emph{flow support} of an action.

The intervention therefore belongs in training, while the policy still  retains meaningful variation to preserve \citep{faria2025sample, bai2025evolve, zhang2026ttt}. The main obstacle is that flow-matching decoders do not directly provide a tractable normalized probability over candidate actions, making standard entropy or distribution-matching regularizers difficult to apply \citep{zhu2026can}. But a flow model does expose something else. Its velocity field has a divergence at every point along an integration path, and integrating that divergence measures how strongly the learned flow contracts toward the action at the end of the path. We call this quantity the \emph{flow support} of an action.

% We introduce \emph{Action Dynamic Sampling} (ADS). At each training step the policy is about to predict action chunk from one base-noise draw; ADS draws $K$ further noise vectors and integrates them into \emph{sibling actions}, the action chunks the policy produces from those draws. 
% Every sibling then carries two numbers. The first number is its flow support, measuring how firmly the policy's flow holds that action. The second is the probability the policy assigns to that same sibling. A KL term pulls the second number toward the first.
We introduce \emph{Action Dynamic Sampling} (ADS). At each training step, in addition to the primary base-noise draw used to produce an action chunk, ADS draws $K$ additional noise vectors and integrates them through the current flow policy to obtain \emph{sibling actions}.
Each sibling is assigned two scores: a flow support score that reflects how strongly the learned flow supports that action, and a local surrogate score that reflects how strongly the current prediction favors it. After normalization over the sibling set, a KL term aligns the local surrogate distribution with the flow support distribution.

% Fine-tuning can make a valid mode rarely sampled while sampled actions in that mode retain substantial flow support.
Fine tuning can make a valid mode increasingly rare while actions that still reach that mode retain substantial flow support.
Probability counts how often a mode gets sampled, while flow support scores individual sampled actions. ADS assigns supported siblings high weight and helps preserve these alternatives. The correction stays tied to what the current policy can reach and vanishes once siblings coincide; it is selective because joint jitter and off-manifold waypoints tend to receive little support \citep{tang2026shifting}. Section~\ref{sec:method} gives the estimator, objective, and gradient.

Our contributions are threefold: (1) we identify proposal contraction as a key constraint on test-time scaling for flow-matching VLA policies, and show that valid modes can become rarely sampled while actions from those modes retain substantial flow support; (2) we introduce ADS, which turns that mismatch into a training signal through a detached finite-set KL term and adds no deployment cost; and (3) we demonstrate consistent gains across simulation and physical-robot tasks, raising oracle best-of-$8$ success on RoboCasa from 57.6\% to 63.0\%, improving fixed verifiers by 2.4 to 4.9 points in simulation and up to 13.0 points in real-world manipulation, with ablations showing that the gain comes from flow support weighting rather than added spread alone.

\section{Problem Formulation}
\label{sec:form}

Let $c=(o,l)$ denote a visual observation and language instruction, and let $a\in\mathbb{R}^{T\times d}$ denote an action chunk.
A VLA policy contains a vision-language encoder $h_\phi=e_\phi(c)$ and an action expert parameterized by $\theta$ \citep{li2026training, black2024pi_0}.
In flow matching, the action expert learns a time-dependent velocity field
\begin{equation}
    \frac{d x_t}{dt}=v_\theta(x_t,t,h_\phi), \qquad t\in[0,1],
\end{equation}
that transports base noise $x_0\sim p_{\mathrm{base}}$ to data actions $x_1=a$, and is trained with
\begin{equation}
\mathcal{L}_{\mathrm{FM}}(\theta,\phi)
=
\mathbb{E}_{t,x_0,a,c}
\left[
\left\|
v_\theta(x_t,t,h_\phi)-u_t(x_0,a)
\right\|_2^2
\right],
\label{eq:fm}
\end{equation}
where $u_t$ is the target interpolation velocity.
Write $\Psi_\theta(z;h_\phi)$ for the endpoint of integrating this field from a
base-noise draw $z$, so that sampling an action means drawing
$z\sim p_{\mathrm{base}}$ and returning
$\mu_\theta(c,z)=\Psi_\theta(z;h_\phi)$.

\paragraph{Proposal contraction.}
For a fixed condition $c$ and demonstration $a$, Eq.~\eqref{eq:fm} supervises trajectories from many base-noise draws toward the same action target.
The quantity this puts at risk is the conditional proposal variance
\begin{equation}
    \mathcal{V}_\theta(c)
    =
    \mathbb{E}_{z}
    \left[
    \left\|
    \mu_\theta(c,z)
    -
    \mathbb{E}_{z'}[\mu_\theta(c,z')]
    \right\|_2^2
    \right],
\label{eq:proposal_variance}
\end{equation}
which measures how much the sampled action still depends on the noise that produced it.
Repeated downstream supervision weakens that dependence and drives $\mathcal{V}_\theta(c)$ toward zero \citep{kim2026libero, peng2026unleashing}.
We call this training trend \emph{proposal contraction}.
Such contraction can accompany improved single-sample fitting while reducing alternatives for verification.

\paragraph{Why contraction caps test-time scaling.}
At test time, a verifier receives $M$ sampled action chunks
\begin{equation}
    \mathcal{C}_M=\{a^{(m)}\}_{m=1}^{M},\qquad
    a^{(m)}\sim\pi_\theta(\cdot\mid c),
\end{equation}
and executes $\arg\max_{a\in\mathcal{C}_M} s(c,a)$ for verifier score $s$.
% What this procedure buys depends on the number of \emph{distinct} candidates, not on the nominal $M$.
The benefit of this procedure depends on the number of distinct candidates rather than the nominal sample count $M$.
Let $\rho$ be the mean off-diagonal correlation of centered candidate embeddings. We summarize candidate similarity using the geometric redundancy index
\begin{equation}
    M_{\mathrm{eff}}=\frac{M}{1+(M-1)\rho}.
\label{eq:meff}
\end{equation}
When $\rho\rightarrow1$, $M_{\mathrm{eff}}\rightarrow1$, indicating highly redundant candidates. This index measures geometric similarity, not the number of statistically independent draws (Appendix~\ref{app:ads_sampling}).
This is the core difficulty.
The budget is spent, the verifier runs $M$ times, and the selection is made among copies of one decision. Redundancy can limit sampling gains even with a strong verifier.
A better verifier cannot repair this, because the modes it would need are never proposed.
Test-time perturbation can increase spread but also displace actions off the action manifold, as observed for Gaussian perturbation in our experiments.

% The gap this leaves is specific.
Contraction is produced during training and limits candidate coverage at test time.
ADS addresses it while $\mathcal{V}_\theta(c)$ is still positive.
\section{Action Dynamic Sampling}
\label{sec:method}

\begin{figure*}
  \centering
  \includegraphics[width=\textwidth]{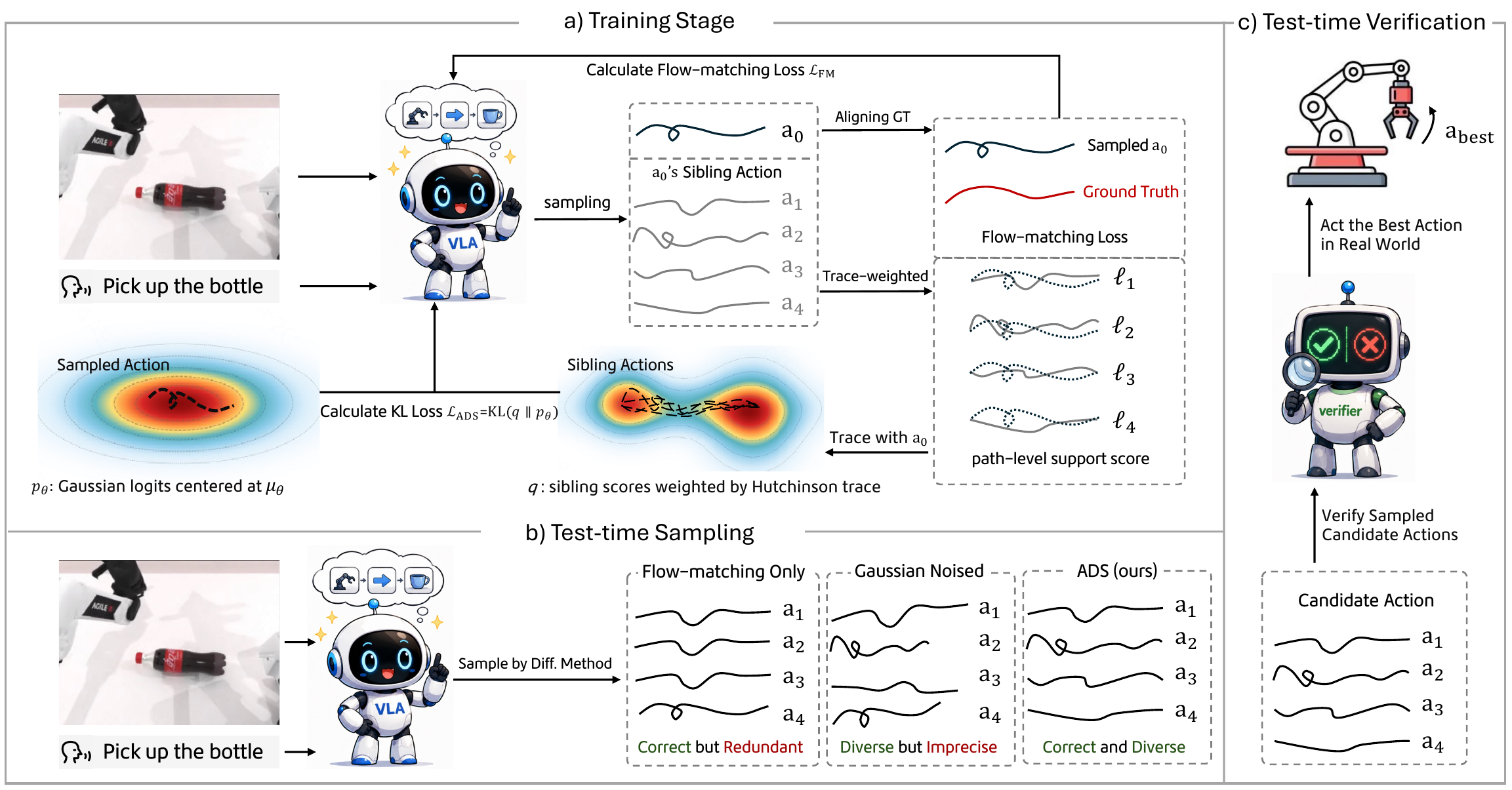}
  \caption{\textbf{Overview of ADS training and test-time verification.}
  During training, detached siblings drawn from the current policy form a finite reference set, a Hutchinson trace estimator weights that set by flow support, and a KL term corrects the policy's local action distribution toward it.
  At test time ADS adds no module: it supplies a broader, policy-aligned candidate set to an unchanged verifier.}
  \label{fig:ads_overview}
\end{figure*}

Fix a condition and sample $K$ sibling action chunks from the policy. For each chunk, ADS evaluates its \emph{flow support}, derived from the divergence of the learned velocity field, and its local Gaussian surrogate score. Fine-tuning can suppress the probability mass of a valid mode while sampled actions in that mode retain flow support. The two scores guide a correction toward supported siblings underweighted by the local surrogate. ADS uses them to construct a finite-set correction that helps preserve such modes, jointly with the standard flow-matching objective.

Section~\ref{sec:flow_support} derives the flow-support score for each sibling, Section~\ref{sec:objective} defines the local probabilities and correction objective, and Section~\ref{sec:correction} analyzes the resulting update.

\subsection{Flow Support over Sibling Actions}
\label{sec:flow_support}
A training step already draws one base-noise vector $z^{(0)}$ and integrates it into the action chunk the supervised loss acts on.
ADS repeats that same procedure $K$ more times, drawing fresh noise and running the same integration \citep{bernat2026addressing}:
\begin{equation}
    z^{(k)}\sim p_{\mathrm{base}},\qquad
    \tilde a^{(k)}=\Psi_\theta(z^{(k)};h_\phi),\qquad
    k=1,\ldots,K,
\label{eq:sibling_set}
\end{equation}
and detaches the resulting endpoints.
We call $\tilde a^{(1:K)}$ the \emph{sibling actions} of the primary draw.
The siblings come only from the policy as it stands at this step, so they show what it would do under different noise.
Sampling from the current policy also limits when the set is useful: contraction pulls the siblings onto the primary chunk, after which they carry no information, so the schedule is a design variable rather than a detail.
ADS is enabled after an initial flow-matching phase, once the action expert is task aligned but $\mathcal{V}_\theta(c)$ in Eq.~\eqref{eq:proposal_variance} has not yet vanished.

The set alone is not yet a training signal, because its members are not equivalent.
A policy midway through contraction produces valid modes, joint jitter, and waypoints that leave the action manifold, and weighting them equally lets the correction chase all three, which costs 6.3 points of success in Section~\ref{sec:experiments}.
The remainder of this subsection scores the siblings so that the correction can prefer among them.

\paragraph{Scoring a sibling.}
The action decoder integrates a velocity field, so the instantaneous change-of-variables formula applies along every sibling path: the log-density evolves at the divergence of that field, the trace of its Jacobian $J_\theta=\partial v_\theta/\partial x_t$.
Integrating from base noise to endpoint therefore yields the log-density the flow itself assigns to that action,
\begin{equation}
    \ell_k
    =
    \log p_{\mathrm{base}}(z^{(k)})
    -
    \int_0^1
    \mathrm{Tr}\left(
    J_\theta(x_t^{(k)},t,h_\phi)
    \right)dt,
\label{eq:path_density}
\end{equation}
We call $\ell_k$ the \emph{flow support} of sibling $k$.
It is large when the field contracts toward that action along the whole path, and small when the sibling lands where the field is spread thin.

Only the trace of $J_\theta$ is needed, which keeps the score affordable: forming the full Jacobian is out of reach for high-dimensional action chunks, but its trace admits an unbiased estimate from $R$ random probes,
\begin{equation}
    \widehat{\mathrm{Tr}}(J_\theta)
    =
    \frac{1}{R}\sum_{r=1}^{R}
    \epsilon_r^\top
    J_\theta
    \epsilon_r,
    \qquad
    \mathbb{E}[\epsilon_r\epsilon_r^\top]=I,
\label{eq:hutchinson}
\end{equation}
evaluated at each ODE discretization point along the sibling trajectory.
Where Jacobian-vector products are inconvenient, a centered finite-difference form of the same estimator costs two velocity evaluations per probe (Appendices~\ref{app_h} and~\ref{app:fair_training_controls}).
The trace term is gated on after training progress $p_{\mathrm{tr}}$; before that, $\ell_k$ is its base-density term alone, and Section~\ref{sec:experiments} ablates the two separately.

\paragraph{From scores to weights.}
A softmax over the $K$ scores turns them into a distribution on the sampled set,
\begin{equation}
    q_k
    =
    \frac{\exp(\ell_k/\tau)}{\sum_{j=1}^{K}\exp(\ell_j/\tau)},
\label{eq:q}
\end{equation}
with temperature $\tau$ setting the sharpness.
Normalization runs over the sampled set alone, so ranking the drawn siblings needs only relative scores.

\subsection{The ADS Objective}
\label{sec:objective}
The correction compares the support-weighted target with a local Gaussian surrogate centered on the primary prediction.
With $\mu_\theta(c,z^{(0)})=\Psi_\theta(z^{(0)};h_\phi)$ the primary endpoint and $\Sigma_\theta$ the fixed diagonal covariance of the baseline sampling perturbation,
\begin{equation}
\begin{aligned}
    r_{\theta,k\mid0}
    &=
    -\tfrac{1}{2}
    \left\|
    \tilde a^{(k)}-\mu_\theta(c,z^{(0)})
    \right\|_{\Sigma_\theta^{-1}}^2,&
    p_{\theta,k\mid0}
    &=
    \frac{\exp(r_{\theta,k\mid0})}
    {\sum_{j=1}^{K}\exp(r_{\theta,j\mid0})}.
\end{aligned}
\label{eq:p}
\end{equation}
The log-determinant of $\Sigma_\theta$ is constant across $k$ and drops out of the softmax.
ADS then minimizes the divergence from the support-weighted target to this distribution,
\begin{equation}
    \mathcal{L}_{\mathrm{ADS}}
    =
    \mathbb{E}_{z^{(0)},z^{(1:K)}}
    \left[
    \mathrm{KL}\!\left(
    \mathrm{sg}[q]
    \,\|\,
    p_{\theta,\cdot\mid0}
    \right)
    \right].
\label{eq:kl}
\end{equation}
Both $q$ and the sibling endpoints are stop-gradient, so the update reaches the action expert only through the primary endpoint $\mu_\theta(c,z^{(0)})$, and the reference is rebuilt from fresh samples at every step.
With $s(p)\in[0,1]$ ramping from zero after progress $p_{\mathrm{ADS}}$, the full objective is
\begin{equation}
    \mathcal{L}
    =
    \mathcal{L}_{\mathrm{FM}}
    +
    s(p)\lambda\mathcal{L}_{\mathrm{ADS}}.
\label{eq:objective}
\end{equation}

\subsection{What the Correction Does}
\label{sec:correction}
Differentiating Eq.~\eqref{eq:kl} through the softmax gives
\begin{equation}
    \frac{\partial \mathcal{L}_{\mathrm{ADS}}}{\partial r_{\theta,k\mid0}}
    =
    p_{\theta,k\mid0}-q_k,
\label{eq:main_kl_grad}
\end{equation}
so the \emph{discrepancy} between the two numbers drives the update.
The logit gradient favors siblings with $q_k>p_{\theta,k\mid0}$.
This encourages an over-localized surrogate to distribute probability across supported siblings when $q$ retains mass on several of them.
We track the size of the correction with the normalized entropy gap
\begin{equation}
    \Delta H
    =
    \frac{H(q)-H(p_{\theta,\cdot\mid0})}{\log K},
\label{eq:rebalancing_diagnostics}
\end{equation}
reported over training in Appendix~\ref{app_i}. Pushed through to the endpoint, the same gradient is a difference of weighted sibling means,
\begin{equation}
\begin{aligned}
    \nabla_{\mu_\theta(c,z^{(0)})}\mathcal{L}_{\mathrm{ADS}}
    =
    \Sigma_\theta^{-1}
    \Big(
    &\textstyle\sum_k p_{\theta,k\mid0}\tilde a^{(k)}
    -
    \textstyle\sum_k q_k\tilde a^{(k)}
    \Big).
\end{aligned}
\label{eq:main_endpoint_gradient}
\end{equation}
The difference of means acts on the denoising dynamics, and two consequences of this form shape the method.
First, if the siblings have already collapsed onto a single action,
\begin{equation}
    \tilde a^{(1)}=\cdots=\tilde a^{(K)}=a^\star
    \;\Longrightarrow\;
    \nabla_{\mu_\theta(c,z^{(0)})}\mathcal{L}_{\mathrm{ADS}}=0,
\label{eq:collapsed_fixed_point}
\end{equation}
since both weighted means equal $a^\star$ regardless of the weights.
This fixed point is what sets the schedule: ADS must act while siblings still disagree.
Second, a first-moment correction over separated modes can point between them. Concentrating $q$ on supported siblings helps reduce invalid inter-mode candidates, as measured in Appendix~\ref{app_i}.

\paragraph{Division of labor.}
$\mathcal{L}_{\mathrm{FM}}$ anchors task correctness; sibling sampling supplies on-policy variation; flow support makes the correction selective; $\lambda$ and $\tau$ set the quality--spread trade-off.
Section~\ref{sec:experiments} measures what each contributes through coverage, rollout validity, and verifier gains.

\section{Experiments}
\label{sec:experiments}

We evaluate ADS across four manipulation benchmarks spanning single-arm and dual-arm embodiments, pairing it with two test-time verifiers that we leave unmodified throughout.
The evaluation addresses four research questions that map one-to-one to the subsections below:

\textbf{Q1:} With no verifier in the loop, does ADS widen the candidate set without giving up proposal quality? (Section~\ref{sec4.1})

\textbf{Q2:} Does a wider candidate set improve fixed test-time verifiers across benchmarks and embodiments? (Section~\ref{sec4.2})

\textbf{Q3:} Which part of the objective produces the gain, and what does it cost to train? (Section~\ref{sec4.3})

\textbf{Q4:} Does the benefit grow with the candidate budget, as alleviating proposal contraction suggests? (Section~\ref{sec4.4})

\paragraph{Setup and metrics.}
We implement ADS in StarVLA with a GR00T-style flow-matching action expert \citep{starvla2026,gr00t2025} and evaluate on LIBERO-Plus, SimplerEnv, RoboCasa, RoboTwin~2.0 \citep{fei2025libero,libero2023,simplerenv2024,robocasa2024,robotwin2025}, and four real world tabletop tasks.
We report success, pairwise proposal RMSE, oracle coverage, valid modes, geometric redundancy index, and training-set replay distances.
Oracle coverage rolls out every candidate, so it upper-bounds a deployed verifier and isolates proposal quality.
Controlled pairs share the backbone, data, horizon, budget, and protocol; Appendices~\ref{app_b}, \ref{app_c}, \ref{app_g} and~\ref{app_h} give details.

\subsection{Proposal Quality and Spread without Verification}
\label{sec4.1}
We first disable verification so only the proposal policy varies.
Table~\ref{table:gaussian_noise_vs_ads} compares flow matching, post-hoc Gaussian perturbation, and ADS at a matched budget.
ADS raises unverified success by 1.5 points on RoboCasa, 2.9 on LIBERO-Plus, and 4.0/1.0 on the two SimplerEnv embodiments, while raising spread from 0.003 to 0.021.
At $\sigma=20\%$, Gaussian noise reaches 0.044 spread but loses 15.1 points on RoboCasa.

\begin{table*}[t]
    \centering
    \caption{\textbf{Gaussian action perturbation and ADS
    without test-time verification.}
    Task performance is success rate (\%).
    Proposal spread is the root mean squared pairwise distance
    among action chunks sampled for the same condition.
    Stronger Gaussian noise buys spread and pays for it in success;
    ADS gains both.}
    \resizebox{0.85\linewidth}{!}{
    \begin{tabular}{l | c c c c || c}
        \toprule
        \textbf{Method}
        & \textbf{RoboCasa}
        & \textbf{LIBERO-Plus}
        & \textbf{WidowX-Bridge}
        & \textbf{Google Robot}
        & \textbf{Proposal spread} \\
        \midrule

        Baseline
        & 54.9
        & 76.5
        & 58.6
        & 69.1
        & 0.003 \\

        Gaussian Noise ($\sigma = 5\%$)
        & 53.3
        & 75.2
        & 58.4
        & 63.2
        & 0.011 \\

        Gaussian Noise ($\sigma = 10\%$)
        & 45.6
        & 73.4
        & 56.3
        & 56.3
        & 0.025 \\

        Gaussian Noise ($\sigma = 20\%$)
        & 39.8
        & 65.9
        & 55.1
        & 50.7
        & 0.044 \\

        \rowcolor{blue!4}
        \tb{ADS}
        & \tb{56.4}
        & \tb{79.4}
        & \tb{62.6}
        & \tb{70.1}
        & 0.021 \\

        \bottomrule
    \end{tabular}
    }

    \label{table:gaussian_noise_vs_ads}
\end{table*}

\begin{figure}[t]
  \centering
  \includegraphics[width=\linewidth]{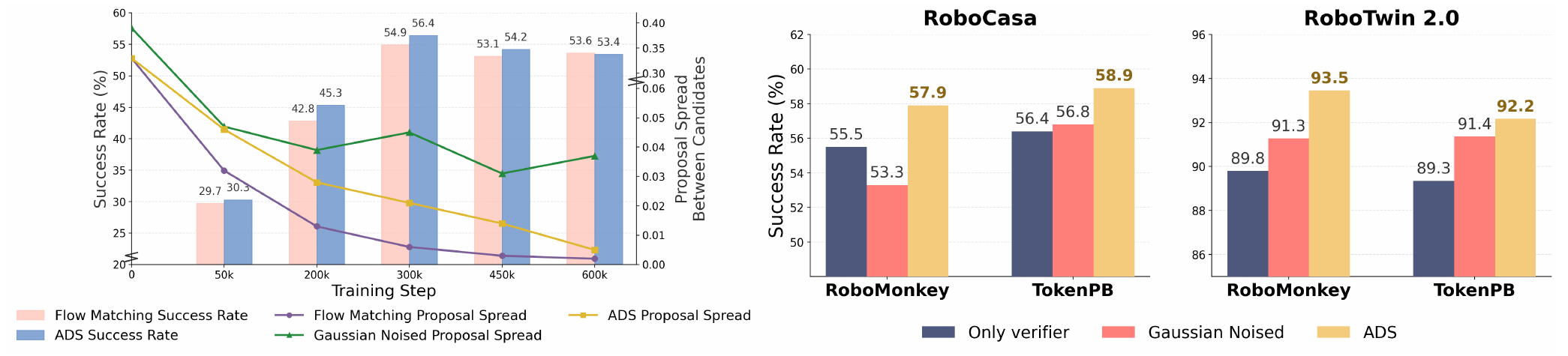}
  \caption{\textbf{Left: Proposal quality and spread over training.} ADS holds a broader candidate set at comparable task success, while Gaussian perturbation is spread out but less policy aligned.
  \textbf{Right: ADS complements test-time verification.} ADS improves both RoboMonkey and TokenPB without changing the verifier.}
  \label{fig:quality_spread}
\end{figure}
\paragraph{Does the wider set stay near valid modes?}
Training-set replay answers this: Table~\ref{table:training_replay} raises $d_{\mathrm{mean}}$ from 0.091 to 0.102 but lowers $d_{\mathrm{best}}$ from 0.085 to 0.032.
ADS therefore spreads samples from one demonstration while placing one candidate much nearer to it; the best distance is the relevant statistic because one demonstration represents only one valid mode.
The signal vanishes after all siblings coincide, motivating future spread-based early stopping.

\begin{table}[t]
\centering
\caption{\textbf{Training-set replay diagnostic on simulation and real-world tasks.}
For each training condition, we sample $M=8$ action chunks and measure the distance from each to the demonstrated action.
$d_{\mathrm{mean}}$ averages over all candidates, while $d_{\mathrm{best}}$ first selects the closest candidate for each condition.}
\label{table:training_replay}
\resizebox{\linewidth}{!}{
\begin{tabular}{l cc cc cc cc}
\toprule
\multirow{2}{*}{\textbf{Metric}}
& \multicolumn{2}{c}{\textbf{Robocasa}}
& \multicolumn{2}{c}{\textbf{RoboTwin 2.0}}
& \multicolumn{2}{c}{\textbf{LIBERO-Plus}}
& \multicolumn{2}{c}{\textbf{Real-World}} \\
\cmidrule(lr){2-3}
\cmidrule(lr){4-5}
\cmidrule(lr){6-7}
\cmidrule(lr){8-9}
& Flow-matching & \cellcolor{blue!4} ADS
& Flow-matching & \cellcolor{blue!4} ADS
& Flow-matching & \cellcolor{blue!4} ADS
& Flow-matching & \cellcolor{blue!4} ADS \\
\midrule

\textbf{Mean distance $d_{\mathrm{mean}}$}
& 0.091 & \cellcolor{blue!4} 0.102
& 0.042 & \cellcolor{blue!4} 0.075
& 0.086 & \cellcolor{blue!4} 0.082
& 0.057 & \cellcolor{blue!4} 0.068 \vspace{1.5px}\\

\textbf{Best distance $d_{\mathrm{best}}$}
& 0.085 & \cellcolor{blue!4} \textbf{0.032}
& 0.036 & \cellcolor{blue!4} \textbf{0.024}
& 0.065 & \cellcolor{blue!4} \textbf{0.027}
& 0.046 & \cellcolor{blue!4} \textbf{0.013} \\
\bottomrule
\end{tabular}
}
\end{table}
\begin{table*}[t]
    \centering
    \caption{\textbf{Proposal-side strategies on
    LIBERO-Plus.}
    The first group contains verification methods with an unmodified
    proposal policy; the second group augments
    the verifier with Gaussian perturbation or ADS.}
    \resizebox{0.92\linewidth}{!}{
    \begin{tabular}{l | c c c c c c c || c}
        \toprule
        \textbf{Method}
        & \textbf{Background}
        & \textbf{Light}
        & \textbf{Language}
        & \textbf{Noise}
        & \textbf{Layout}
        & \textbf{Robot Init.}
        & \textbf{Camera}
        & \textbf{Overall} \\
        \midrule

        RoboMonkey
        & 94.1
        & 93.2
        & \underline{91.7}
        & 77.8
        & 82.7
        & 57.0
        & 56.5
        & 77.6 \\

        TokenPB
        & 96.7
        & 95.6
        & 86.7
        & 79.1
        & 83.1
        & 55.2
        & 52.2
        & 76.9 \\

        \midrule

        RoboMonkey + Gaussian Noise
        & 96.9
        & 94.1
        & 91.2
        & 80.8
        & 85.5
        & 56.9
        & 58.6
        & 79.4 \\

        \rowcolor{blue!4}
        RoboMonkey + ADS
        & \underline{97.9}
        & \underline{96.7}
        & \tb{93.9}
        & \underline{81.6}
        & \underline{88.1}
        & \tb{58.5}
        & \tb{62.5}
        & \underline{81.3} \\

        \rowcolor{blue!4}
        TokenPB + ADS
        & \tb{99.0}
        & \tb{97.6}
        & 90.6
        & \tb{84.7}
        & \tb{90.2}
        & \underline{57.8}
        & \underline{61.4}
        & \tb{81.6} \\

        \bottomrule
    \end{tabular}
    }

    \label{table:libero_plus_ads}
\end{table*}

\begin{table*}[t]
    \centering
    \caption{\textbf{Proposal-side strategies on
    SimplerEnv.}
    The first group contains methods with an unmodified proposal
    policy; the second group introduces Gaussian
    perturbation or ADS.}
    \resizebox{\linewidth}{!}{
    \begin{tabular}{l | c c c c c || c c c c c}
        \toprule
        & \multicolumn{5}{c||}{\textbf{WidowX-Bridge (SR)}}
        & \multicolumn{5}{c}{\textbf{Google Robot (SR)}} \\

        \textbf{Method}
        & \textbf{Stack Cube}
        & \textbf{Carrot on Plate}
        & \textbf{Eggplant in Basket}
        & \textbf{Spoon on Towel}
        & \textbf{Avg.}
        & \textbf{Drawer}
        & \textbf{Move Near}
        & \textbf{Pick Coke}
        & \textbf{Put in Drawer}
        & \textbf{Avg.} \\
        \midrule

        RoboMonkey
        & 38.2
        & 64.7
        & 86.7
        & 72.9
        & 65.6
        & --
        & --
        & --
        & --
        & -- \\

        TokenPB
        & \underline{46.9}
        & 70.8
        & \tb{100.0}
        & \tb{85.4}
        & \underline{75.8}
        & \underline{57.1}
        & \underline{83.7}
        & \underline{86.8}
        & 50.3
        & \underline{76.4} \\

        \midrule

        Gaussian Noise + TokenPB
        & 26.4
        & 65.5
        & 89.6
        & 65.8
        & 61.8
        & 55.8
        & 76.4
        & 82.5
        & \tb{58.6}
        & 73.3 \\

        \rowcolor{blue!4}
        RoboMonkey + ADS
        & 46.5
        & \underline{71.9}
        & 87.5
        & 76.0
        & 70.5
        & --
        & --
        & --
        & --
        & -- \\

        \rowcolor{blue!4}
        TokenPB + ADS
        & \tb{58.3}
        & \tb{77.1}
        & \underline{97.9}
        & \underline{83.3}
        & \tb{79.2}
        & \tb{59.6}
        & \tb{93.5}
        & \tb{89.4}
        & \underline{57.1}
        & \tb{79.4} \\

        \bottomrule
    \end{tabular}
    }
    \label{table:widowx_google_robot_ads}
\end{table*}

\subsection{Coupling ADS with Test-Time Verifiers}
\label{sec4.2}
Q1 establishes a better set; Q2 asks what fixed verifiers do with it. We hold verifier weights and budget fixed while changing proposal training.
On LIBERO-Plus (Table~\ref{table:libero_plus_ads}), ADS raises RoboMonkey/TokenPB success from 77.6/76.9\% to 81.3/81.6\%.
On SimplerEnv (Table~\ref{table:widowx_google_robot_ads}), it raises RoboMonkey from 65.6\% to 70.5\% on WidowX-Bridge and TokenPB by 3.4/3.0 points across its two embodiments.
RoboMonkey also rises from 55.5\% to 57.9\% on RoboCasa and 89.8\% to 93.5\% on RoboTwin~2.0.
Gaussian perturbation is inconsistent; ADS improves every verifier--benchmark pair by 2.4--4.9 points, showing that fixed verifiers exploit its more valid candidates.
\begin{table}[t]
    \centering
    \caption{\textbf{Component ablation on RoboCasa without verifier.}
    Each row adds one element of the ADS objective to the one above it, averaged over five seeds.
    Success is single-sample; proposal spread is measured among candidates drawn for the same condition.}
    \resizebox{0.75\linewidth}{!}{
    \begin{tabular}{l c c}
        \toprule
        \textbf{Training objective}
        & \textbf{Success (\%)}
        & \textbf{Proposal spread} \\
        \midrule
        $\mathcal{L}_{\mathrm{FM}}$
        & 54.9
        & 0.003 \\
        $+\,$uniform siblings ($q_k=1/K$)
        & 48.6
        & 0.029 \\
        $+\,$base-density weighting ($\ell_k=\log p_{\mathrm{base}}(z^{(k)})$)
        & 53.2
        & 0.023 \\
        \rowcolor{blue!4}
        $+\,$flow support (full ADS)
        & 56.4
        & 0.021 \\
        \bottomrule
    \end{tabular}
    }
    \label{table:ads_component_ablation}
\end{table}

\begin{table}[!b]
    \centering
    \caption{\textbf{Candidate scaling on RoboCasa with RoboMonkey.}
    Each block reports one property of the same $M$ candidates, using the metrics defined in the experimental setup.
    Oracle coverage and verifier success differ only in who selects the executed chunk.
    The ADS advantage widens with $M$ before saturating.}
    \resizebox{0.76\linewidth}{!}{
    \begin{tabular}{c | c c | c c | c c | c c}
        \toprule
        & \multicolumn{2}{c|}{\textbf{Geometric redundancy}}
        & \multicolumn{2}{c|}{\textbf{Valid modes}}
        & \multicolumn{2}{c|}{\textbf{Oracle coverage (\%)}}
        & \multicolumn{2}{c}{\textbf{Verifier success (\%)}} \\
        \textbf{Candidates $M$}
        & \textbf{FM} & \textbf{ADS}
        & \textbf{FM} & \textbf{ADS}
        & \textbf{FM} & \textbf{ADS}
        & \textbf{FM} & \textbf{ADS} \\
        \midrule
        1  & 1.00 & 1.00 & 1.00 & 1.00 & 54.9 & 56.4 & 54.9 & 56.4 \\
        2  & 1.04 & 1.35 & 1.02 & 1.25 & 55.9 & 58.4 & 55.1 & 56.9 \\
        4  & 1.06 & 1.62 & 1.05 & 1.63 & 56.8 & 60.7 & 55.3 & 57.5 \\
        \rowcolor{blue!4}
        8  & 1.08 & 2.05 & 1.08 & 2.30 & 57.6 & 63.0 & 55.5 & 57.9 \\
        16 & 1.09 & 2.30 & 1.10 & 2.85 & 58.0 & 64.2 & 55.6 & 58.1 \\
        \bottomrule
    \end{tabular}
    }

    \label{table:test_time_scaling}
\end{table}

\begin{figure}[t]
  \centering
  \includegraphics[width=\linewidth]{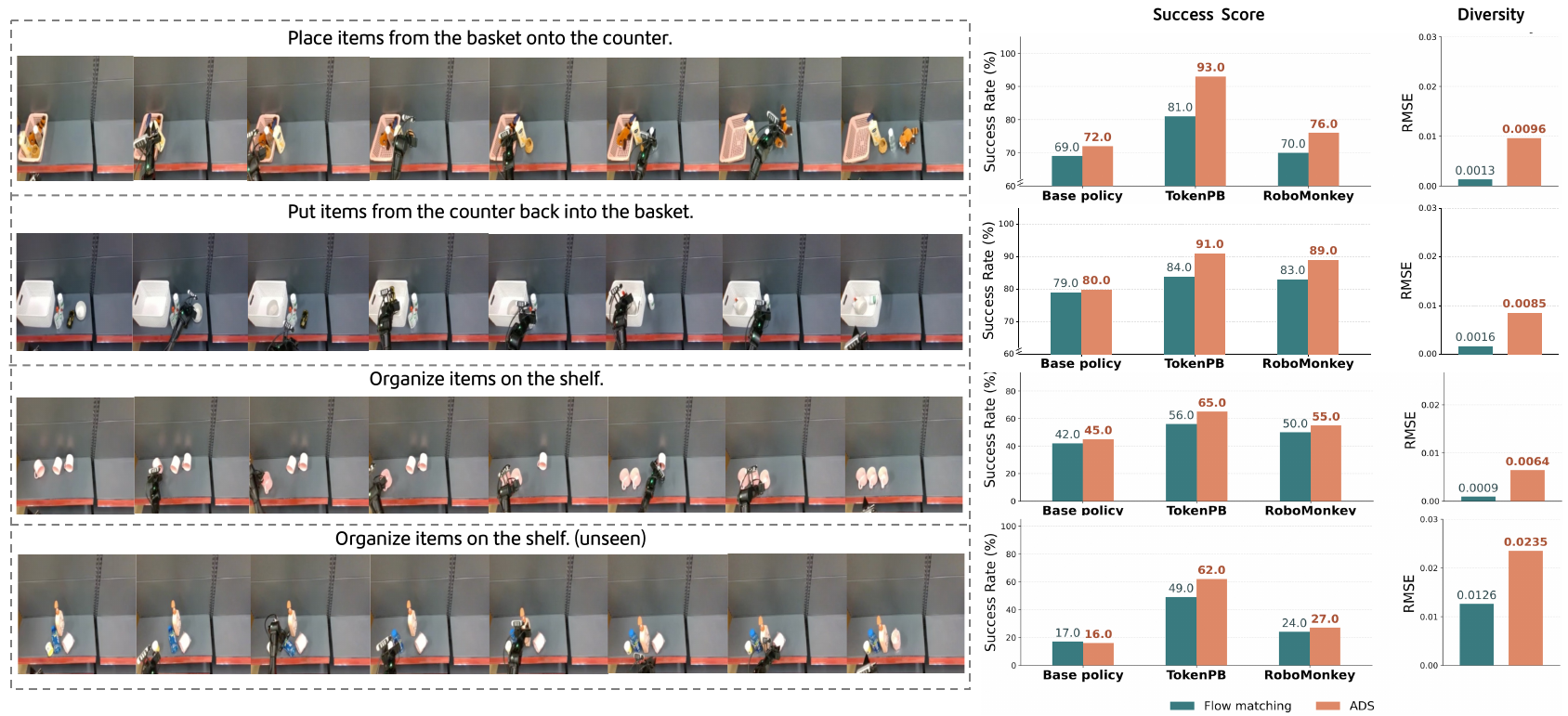}
  \caption{\textbf{Real-world four-object manipulation results.} Left: representative execution sequences. Middle: normalized success scores for unverified, TokenPB-selected, and RoboMonkey-selected policies. Right: proposal diversity measured by pairwise action RMSE. ADS broadens the proposal set and improves both fixed verifiers, including on the unseen shelf task.}
  \label{fig:real_world}
\end{figure}

\paragraph{Real-world evaluation.}
We evaluate four four-object manipulation tasks in a mock supermarket environment:
\emph{Place items from the basket onto the counter},
\emph{Put items from the counter back into the basket},
\emph{Organize items on the shelf}, and
\emph{Organize items on the shelf (unseen)}.
Across these tasks, ADS increases the mean pairwise action RMSE from $0.0041$ to $0.0120$ ($2.9\times$), indicating a substantially broader proposal set.
When coupled with test-time verification, this additional diversity yields consistent improvements across all tasks: ADS improves the average success score by $10.3$ points with TokenPB and by $5.0$ points with RoboMonkey.
The gains persist on the unseen shelf-organization task, indicating that ADS provides verifiers with a broader set of executable alternatives rather than unstructured action noise.
Figure~\ref{fig:real_world} reports the per-task results; Appendix~\ref{app:real_world_scoring} gives the setup, scoring, and protocol.

\subsection{Which Component Pays, and What It Costs}
\label{sec4.3}
The gains above leave two questions open: which ingredient of the ADS objective drives them, and what the resulting training overhead is.
This subsection answers both, first isolating the contribution of each objective component and then quantifying the training-only cost.
\paragraph{Objective components.}
Table~\ref{table:ads_component_ablation} adds sibling resampling and flow-support weighting one at a time.
Uniform siblings maximize spread but drop success from 54.9\% to 48.6\%, because they weight jitter and unsupported waypoints like valid modes.
Base-density weighting recovers to 53.2\%; path-integrated support reaches 56.4\% at 0.021 spread.
The trace therefore judges which variation to preserve, rather than merely increasing it. We report equal-FLOPs FM, early-checkpoint, and entropy-regularization controls with matched training and verification protocols in Appendix~\ref{app:fair_training_controls}. Appendix~\ref{app_i} extends this analysis with a one-at-a-time sensitivity study over $K$, $R$, $\tau$, and $\lambda_{\mathrm{ADS}}$, showing that the default configuration lies in a broad, stable operating region.

\paragraph{Training cost.}
ADS acts entirely at training time: the deployed policy and verifier are the baseline's.
Two hyperparameters set the added cost, the number of siblings $K$ integrated per update and the number of Hutchinson probes $R$ used per trace estimate, and every ADS evaluation scales with their product.
Table~\ref{table:ads_efficiency} reports profiler-measured training FLOPs, step time, peak memory, and deployment FLOPs as both are varied.
The default $K=8,R=8$ costs 3.37$\times$ the training FLOPs of flow matching; because sibling integration and trace probes batch, the wall-time increase is 1.28$\times$, and deployment cost is identical by construction.
Halving both gives a cheaper operating point: $K=4,R=4$ costs 1.67$\times$ FLOPs and reaches 56.1\% against the default's 56.4\%, and Appendix~G shows the trace estimator saturating by $R=4$. the full accounting is in Appendix~\ref{app:training_compute}.

\subsection{Candidate-Budget Scaling}
\label{sec4.4}
Section~\ref{sec:form} argues that contraction caps test-time scaling by driving $M_{\mathrm{eff}}\rightarrow1$.
Table~\ref{table:test_time_scaling} tests the converse: if ADS removes contraction, its advantage should widen with $M$.
The flow-matching baseline holds $M_{\mathrm{eff}}$ near one at every budget: at $M=16$ it draws sixteen candidates and receives the value of 1.09 independent ones, while ADS reaches 2.30.
Valid modes move the same way: at $M=8$ the baseline occupies 1.08 and ADS occupies 2.30, so ADS hands the verifier roughly two working ways to finish the task instead of one.
Oracle coverage widens in step, from a 1.5-point gap at $M=1$ to 6.2 points at $M=16$, so the added spread is executable and a larger budget reaches more of it.
Verifier success follows the same ordering at a lower level, and the gap to oracle coverage is what the selector misses.
Those five points are already in the set and go unselected, so removing proposal contraction moves the binding constraint to the selector.

\section{Related Work}

\paragraph{Flow and diffusion action models.}
Diffusion Policy showed that denoising generative models can represent multimodal visuomotor actions \citep{chi2023diffusionpolicy}, and flow matching learns a velocity field that transports a simple base distribution to data \citep{lipman2023flowmatching,tong2023conditional}.
These models make good action decoders, although their training objectives do not require explicit likelihood evaluation.
Continuous normalizing flows recover likelihoods through the instantaneous change-of-variables formula, whose divergence term admits a Hutchinson estimator \citep{chen2018node,grathwohl2019ffjord,hutchinson1989}.
ADS uses the resulting log-density estimate as a relative ranking over a sampled finite set, which is all a normalized-within-set weighting requires.
Where entropy maximization and global distribution matching specify a target shape for the policy, ADS specifies a reference the policy is corrected toward, and the direction of each correction follows from the discrepancy between the two.

\paragraph{Test-time verification.}
Verifier-based VLA scaling samples candidate actions and selects one with an external or learned verifier \citep{kwok2026scaling, sun2026pre, zhao2026verispace, ali2025eve}.
RoboMonkey shows that sampling plus verification improves VLA execution when the candidate set contains valid modes to choose among \citep{kwok2025robomonkey}.
That precondition is where ADS intervenes: rather than building a stronger selector, it trains the proposal policy so the selector has something to select from.
We also pair ADS with TokenPB \citep{zhao2026action}, a self-verify method evaluated under the same test protocol, which combine the sampler and the verifier into the same checkpoint.

\section{Conclusion}

Sampling more actions helps only if the samples differ.
Fine-tuning a flow-matching VLA policy drives its candidates together, but valid alternatives can retain flow support as they become rarely sampled, and Action Dynamic Sampling helps preserve these alternatives during training.
Across four simulation benchmarks and real world environment, ADS widens the candidate set and raises success at the same time, and both verifiers we pair it with convert that wider set into higher selected success.
The cost falls entirely on training: the deployed policy and verifier are the baseline's, and the advantage grows across the evaluated candidate budgets, as ADS supplies more distinct candidates while flow matching remains highly redundant.

\subsection*{AI use statement}
In this work, we used generative AI tools to assist with writing and polishing the manuscript, checking the code, and developing and verifying the proof of the diagnostic construction. We have not used generative AI tools for any other tasks requiring disclosure, and the remaining required disclosure categories are not applicable to this work. All AI-assisted work was reviewed and confirmed by the main authors against the underlying code, experiments, and mathematical derivations. We take full responsibility for the final content of this work, including all text, claims, code, and artifacts produced with the aid of generative AI.
\subsection*{Reproducibility Statement}

We provide the VLA architectures, experimental settings, and statistical
evaluation protocol in Appendix~\ref{app_c}. Implementation details and
hyperparameters are documented in Appendix~\ref{app_h}, with supplementary
ablation results and statistical diagnostics in Appendix~\ref{app_i}.
Appendix~\ref{app_g} presents the training algorithm through a detailed
flowchart and pseudocode, and the subsequent theoretical appendices
analyze independent sampling, geometric redundancy, flow-support weighting,
and the relative rebalancing induced by the ADS objective. For the
real-world experiments, Appendix~\ref{app:real_world_scoring} specifies
the equipment and scoring criteria and includes an overview of the
experimental setup. All training and inference code is included in
the supplementary materials. Formal assumptions, proofs, and controlled diagnostics are provided in Appendices~\ref{app:ads_theory_scope} and~\ref{app:ads_synthetic}; sampling and evaluation conventions are detailed in Appendices~\ref{app:ads_sampling} and~\ref{app:ads_reporting}.
\bibliographystyle{iclr2027_conference}
\bibliography{references}
\clearpage
\appendix
\providecommand{\adsresult}[1]{\if\relax\detokenize{#1}\relax\underline{\hspace{0.68cm}}\else#1\fi}

\section{Method and Notation Recap}

This supplement specifies the objective, evaluation conventions, and the assumptions under which the finite-set correction can be interpreted.
For condition $c=(o,l)$, the StarVLA action expert defines a velocity field $v_\theta(x_t,t,h_\phi)$ and is trained with
\begin{equation}
\mathcal L_{\mathrm{FM}}
=
\mathbb E_{t,x_0,a,c}
\left[
\left\|v_\theta(x_t,t,h_\phi)-u_t(x_0,a)\right\|_2^2
\right].
\label{eq:supp_fm}
\end{equation}
ADS samples a primary endpoint $\mu_\theta(c,z^{(0)})$ and detached siblings
\begin{equation}
\begin{aligned}
z^{(k)}&\sim p_{\mathrm{base}}, \qquad k=1,\ldots,K,\\
\tilde a^{(k)}&=\Psi_\theta(z^{(k)};\mathrm{sg}[h_\phi]).
\end{aligned}
\end{equation}
The finite-difference Hutchinson estimator used along sibling trajectories is
\begin{equation}
\begin{aligned}
\widehat{\mathrm{Tr}}_\delta(J_\theta)
&=
\frac{1}{R}\sum_{r=1}^{R}
\epsilon_r^\top\frac{\Delta v_{\theta,r}}{2\delta},\\
\Delta v_{\theta,r}
&=
v_\theta(x_t+\delta\epsilon_r,t,h_\phi)\\
&\quad-v_\theta(x_t-\delta\epsilon_r,t,h_\phi).
\end{aligned}
\label{eq:supp_finite_trace}
\end{equation}
With normalized training progress $p$ and trace activation $\gamma(p)=\mathbf 1[p\ge p_{\mathrm{tr}}]$, the local support score and finite-set target are
\begin{align}
\ell_k(p)
&=
\log p_{\mathrm{base}}(z^{(k)})
-\gamma(p)\int_0^1\mathrm{Tr}(J_\theta(x_t^{(k)},t,h_\phi))dt,
\\
q_k(p)
&=
\frac{\exp(\ell_k(p)/\tau)}{\sum_j\exp(\ell_j(p)/\tau)}.
\label{eq:supp_q}
\end{align}
For fixed diagonal $\Sigma$, the primary-conditioned surrogate is
\begin{equation}
p_{\theta,k\mid0}
=
\frac{
\exp\!\left(-\frac12\|\tilde a^{(k)}-\mu_\theta(c,z^{(0)})\|_{\Sigma^{-1}}^2\right)
}{
\sum_j\exp\!\left(-\frac12\|\tilde a^{(j)}-\mu_\theta(c,z^{(0)})\|_{\Sigma^{-1}}^2\right)
}.
\label{eq:supp_p}
\end{equation}
The training objective is
\begin{equation}
\mathcal L
=
\mathcal L_{\mathrm{FM}}
+s(p)\lambda\,
\mathbb E_{z^{(0)},z^{(1:K)}}
\left[
\mathrm{KL}\!\left(\mathrm{sg}[q(p)]\,\|\,p_{\theta,\cdot\mid0}\right)
\right].
\label{eq:supp_kl}
\end{equation}
Both siblings and $q$ are detached, so the update reaches the action expert only through the primary endpoint and its surrogate logits. The conditional flow density, the probability mass of a region of action space, and the finite-set surrogate in Eq.~\eqref{eq:supp_p} are different quantities; Appendix~\ref{app:ads_theory_scope} relates them without identifying one with another.

\section{Real-World Evaluation Protocol and Scoring Details}
\label{app:real_world_scoring}
\label{app_b}

\begin{figure}[t]
    \centering
    \includegraphics[width=0.5\linewidth]{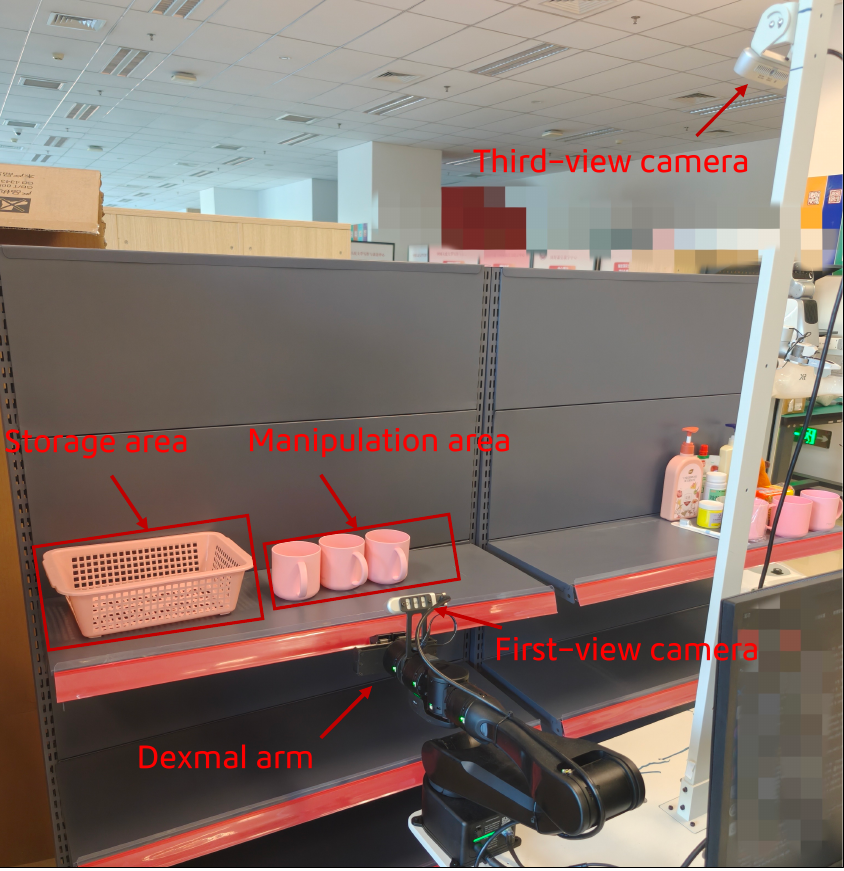}
    \caption{Real-world evaluation setup. The workspace consists of a
    \emph{storage area}, a \emph{manipulation area}, a first-person-view
    camera, a third-person-view camera, and a Dexmal robotic arm.}
    \label{fig:real_demo}
\end{figure}

As illustrated in Figure~\ref{fig:real_demo}, our real-world setup contains
a storage area, a manipulation area, a first-person-view camera, a
third-person-view camera, and a Dexmal robotic arm. We evaluate four
four-object manipulation tasks: \textit{Place items from the basket onto the
counter}; \textit{Put items from the counter back into the basket};
\textit{Organize items on the shelf}; and \textit{Organize items on the shelf
(unseen)}. Here, the basket serves as the storage area, while the counter
serves as the manipulation area.

For each rollout, we report a normalized task score $S \in [0,1]$ that
captures both partial progress and complete task execution. Real-world percentages
in the main paper summarize the mean normalized task score, $100\,\mathbb E[S]$.
This quantity is distinct from the percentage of rollouts completing all object-level
subgoals, $100\Pr(c=1)$, where $c$ is the binary completion indicator defined below.
Appendix~\ref{app:ads_reporting} gives the separate estimators and their relationship.
Each task contains $N=4$ target objects. In addition to the per-object
partial score, we assign a global completion bonus, so that fully completing
all four object-level subgoals is necessary to receive a score of $1.0$.

\paragraph{Transfer between the storage area and the manipulation area.}
We use the same scoring rule for transferring objects from the basket to the
counter and for transferring them from the counter back to the basket. Let
$A$ be the source area and $B$ be the destination area, and let
$s_i^{A\rightarrow B}$ denote the score of target object $i$:
\begin{equation}
s_i^{A\rightarrow B} =
\begin{cases}
0.0, & \text{if object $i$ is not successfully grasped and lifted from $A$},\\
0.8, & \text{if object $i$ is lifted from $A$ but does not finish stably in $B$},\\
1.0, & \text{if object $i$ finishes stably within the designated region of $B$}.
\end{cases}
\end{equation}
We further define the binary completion indicator
$c^{A\rightarrow B} \in \{0,1\}$, where $c^{A\rightarrow B}=1$ only if all
four target objects are stably placed in the designated destination area at
the end of the rollout, with no target object left in the source area or
dropped outside the workspace. The transfer score is
\begin{equation}
S_{A\rightarrow B}
=
\frac{1}{N+1}
\left(
\sum_{i=1}^{N} s_i^{A\rightarrow B}
+
c^{A\rightarrow B}
\right),
\qquad N=4.
\label{eq:transfer_score}
\end{equation}
Accordingly, \textit{Place items from the basket onto the counter} uses
$A=\mathcal{R}_{\mathrm{storage}}$ and
$B=\mathcal{R}_{\mathrm{manipulation}}$, whereas \textit{Put items from the
counter back into the basket} uses
$A=\mathcal{R}_{\mathrm{manipulation}}$ and
$B=\mathcal{R}_{\mathrm{storage}}$. The first four terms in
Equation~\eqref{eq:transfer_score} contribute $80\%$ of the total score,
while the final $20\%$ rewards complete execution of all four transfers.

\paragraph{Shelf organization.}
For the shelf tasks, let $\mathcal{G}=\{g_i\}_{i=1}^{N}$ denote the
predefined target shelf arrangement, where $g_i$ specifies the desired
shelf position of object $i$. We assign the following per-object score:
\begin{equation}
q_i =
\begin{cases}
0.0, & \text{if object $i$ is not successfully grasped and lifted},\\
0.5, & \text{if object $i$ is grasped but does not finish stably on the shelf},\\
0.8, & \text{if object $i$ finishes stably on the shelf but outside $g_i$},\\
1.0, & \text{if object $i$ finishes stably at its designated target position $g_i$}.
\end{cases}
\end{equation}
Let $c_{\mathrm{shelf}}\in\{0,1\}$ be a binary completion indicator.
$c_{\mathrm{shelf}}=1$ only when all four objects are stably arranged
according to the target configuration $\mathcal{G}$, including the required
object-to-position assignments and relative ordering, with no object left
outside the shelf or dropped during execution. The shelf-organization score is
\begin{equation}
S_{\mathrm{shelf}}
=
\frac{1}{N+1}
\left(
\sum_{i=1}^{N} q_i + c_{\mathrm{shelf}}
\right),
\qquad N=4.
\label{eq:shelf_score}
\end{equation}

The \textit{Organize items on the shelf} task and the \textit{Organize items
on the shelf (unseen)} task use the identical scoring function in
Equation~\eqref{eq:shelf_score}. The unseen setting differs only in that it
uses held-out task instances; no scoring criterion is relaxed for the unseen
evaluation.

\section{Experimental Details}
\label{app_c}

\paragraph{Policy architecture.}
All policies use StarVLA with a GR00T-style vision--language backbone and flow-matching action expert.
ADS is applied only to the action expert: the ordinary flow-matching loss updates the full VLA, while the ADS condition, sibling endpoints, and weighted finite-set target are detached from the vision--language encoder.
The visual and language representation, action parameterization, and deployment-time policy are identical between the controlled baseline and ADS comparisons.

\begin{table*}[t]
\caption{VLA architecture used in all controlled experiments.
The baseline and ADS share every component; ADS changes only the training objective applied to the action expert.}
\centering
\resizebox{0.92\linewidth}{!}{
\begin{tabular}{l l l}
\toprule
\textbf{Component} & \textbf{Configuration} & \textbf{Parameters updated by ADS KL} \\
\midrule
Vision-language backbone & StarVLA base, 2.1B parameters, 224$\times$224 RGB input & No \\
Language context & 128 tokens, task instruction and embodiment tokens & No \\
Action expert & GR00T-style flow transformer, width 1024, 16 blocks, 16 heads & Yes \\
Action representation & normalized joint/EEF action chunks with gripper state & Primary endpoint only \\
Condition interface & 32 projected vision-language tokens & Detached \\
Numerical precision & bfloat16 backbone, float16 trace, float32 optimizer states & -- \\
\bottomrule
\end{tabular}}
\label{table:supp_architecture}
\end{table*}

\paragraph{Benchmarks.}
We use four simulation suites: LIBERO-Plus evaluates visual and language shifts; SimplerEnv reports WidowX-Bridge and Google Robot separately; RoboCasa covers diverse household manipulation; and RoboTwin~2.0 evaluates dual-arm manipulation.
We additionally evaluate four physical-robot tasks with four target objects each: basket-to-counter transfer, counter-to-basket transfer, shelf organization, and unseen shelf organization.
SimplerEnv follows the official task-specific evaluation configurations, whose episode counts differ across tasks. Its aggregate success rate is computed by pooling successful and evaluated episodes, equivalently by weighting each task success rate by its evaluation count; it is not the unweighted mean of the displayed task percentages. WidowX-Bridge and Google Robot are aggregated separately. Appendix~\ref{app:ads_reporting} gives the estimator and distinguishes it from an unweighted group mean. All held-out episodes are excluded from checkpoint selection.
Training and validation split by full demonstration trajectory.

\begin{table*}[t]
\caption{Dataset splits and evaluation budgets.
SimplerEnv is one suite but its WidowX-Bridge and Google Robot embodiments are kept as separate rows and are never averaged together. Its evaluation counts follow the official task-specific configurations.}
\centering
\resizebox{0.98\linewidth}{!}{
\begin{tabular}{l c c c c c c}
\toprule
\textbf{Suite / embodiment} & \textbf{Tasks} & \textbf{Train demos/task} & \textbf{Train/val} & \textbf{Action dim. / horizon} & \textbf{Updates} & \textbf{Eval. episodes/task/seed} \\
\midrule
LIBERO-Plus & 40 & 42 & - & 7 / 16 & 40k & 50 per shift \\
SimplerEnv: WidowX-Bridge & 4 & --- & 99\% / 1\% & 7 / 8 & 50k & task-dependent \\
SimplerEnv: Google Robot & 4 & --- & 99\% / 1\% & 7 / 8 & 50k & task-dependent \\
RoboCasa & 24 & 100 & 99\% / 1\% & 7 / 16 & 60k & 50 \\
RoboTwin~2.0 & 50 & 550 & 95\% / 5\% & 14 / 16 & 60k & 50 \\
\bottomrule
\end{tabular}}
\label{table:supp_benchmark_protocol}
\end{table*}

\paragraph{Observation, action, and augmentation processing.}
RGB observations are resized to $224\times224$ after a random resized crop with scale in $[0.9,1.0]$ during training; color jitter has magnitude 0.1 and is disabled at evaluation.
Robot state and continuous action dimensions are standardized with training-set means and standard deviations, then clipped to $[-5,5]$ before entering the action expert.
Binary gripper commands remain unstandardized.
At evaluation, the policy executes the first four actions of each predicted chunk and replans, except in SimplerEnv where the official control frequency executes the first two actions.

\paragraph{Proposal and verifier protocol.}
Proposal-only comparisons disable the verifier and sample candidates from the same checkpoint and condition; Gaussian baselines perturb a predicted chunk, whereas ADS samples from the ADS-trained flow.
Within each controlled verifier pair, candidate count, weights, score, and selection rule are fixed.
The default budget is $M=8$; RoboMonkey and TokenPB are frozen and are not retrained on ADS candidates.
Candidates are scored independently, ties use the lowest index, and the selected chunk uses the same receding-horizon controller as proposal-only execution.
Physical-robot trials randomize the four-object initial configuration; all six method configurations use matched task instances and the same scoring rule in Appendix~\ref{app:real_world_scoring}.

\paragraph{Optimization and model selection.}
The released StarVLA initialization is used for every seed.
We select hyperparameters on the validation split by single-sample success; differences smaller than 0.2 percentage points are broken in favor of lower measured training FLOPs.
The final checkpoint is selected once per seed using validation success and then evaluated on all benchmark variants without per-variant tuning.

\begin{table*}[t]
\caption{Optimization, sampling, and ADS hyperparameters.
The final value is selected by validation single-sample success, with lower training FLOPs used to break differences below 0.2 points.}
\centering
\resizebox{0.96\linewidth}{!}{
\begin{tabular}{l c c || l c c}
\toprule
\textbf{Parameter} & \textbf{Final value} & \textbf{Values evaluated} & \textbf{Parameter} & \textbf{Final value} & \textbf{Values evaluated} \\
\midrule
Optimizer & AdamW & AdamW & Batch size & 64 & 32, 64 \\
Action-expert LR & $1\!\times\!10^{-4}$ & $5\!\times\!10^{-5}$, $10^{-4}$ & Backbone LR & $2\!\times\!10^{-5}$ & $10^{-5}$, $2\!\times\!10^{-5}$ \\
Weight decay & 0.01 & 0.00, 0.01 & Gradient clip & 1.0 & 0.5, 1.0 \\
Warmup updates & 500 & 0, 500 & LR schedule & cosine & constant, cosine \\
Sibling count $K$ & 8 & 2, 4, 8, 16 & Trace probes $R$ & 8 & 1, 2, 4, 8 \\
Temperature $\tau$ & 1.0 & 0.25, 0.5, 1.0, 2.0 & ADS weight $\lambda$ & $10^{-3}$ & $10^{-4}$, $3\!\times\!10^{-4}$, $10^{-3}$, $3\!\times\!10^{-3}$ \\
Gaussian scale $\sigma$ & 0.10 & 0.05, 0.10, 0.20 & Finite-difference $\delta$ & $10^{-4}$ & $10^{-5}$, $10^{-4}$, $10^{-3}$ \\
ADS start $p_{\mathrm{ADS}}$ & 0.50 & 0.30, 0.50, 0.70 & Trace start $p_{\mathrm{tr}}$ & 0.60 & 0.0, 0.3, 0.6, 0.8 \\
ODE steps & 4 & 4, 8 & Candidates $M$ & 8 & 1, 2, 4, 8, 16 \\
Training seeds & 0--4 & five independent seeds & Probe distribution & Rademacher & Gaussian, Rademacher \\
\bottomrule
\end{tabular}}
\label{table:supp_optimization}
\end{table*}

\label{data_seed}
\paragraph{Compute environment.}
Training uses eight NVIDIA H100 80GB GPUs on one node with two AMD EPYC 9654 CPUs and 1TB system memory.
The software stack is Ubuntu 22.04, CUDA 12.1, cuDNN 9.1, Python 3.10, PyTorch 2.3, and NCCL 2.20.
The backbone uses bfloat16 activations, the trace branch uses float16, and optimizer states are maintained in float32.
Profiler measurements exclude dataloader startup and include forward, backward, optimizer, sibling integration, and trace-probe operations.

\paragraph{Statistical protocol.}
Controlled policy pairs use identical task instances, object placements, simulator seeds, and language variants.
Principal comparisons use five training seeds and a hierarchical bootstrap that resamples training seeds, tasks, and then episodes within each task.
Proposal-spread statistics compare candidates conditioned on the same observation and instruction and are additionally stratified by successful and failed episodes where reported.
We use 10,000 bootstrap replicates.
The reported two-sided $p$-value is $2\min\{\Pr(\Delta^*\le0),\Pr(\Delta^*\ge0)\}$ for paired bootstrap differences $\Delta^*$, clipped below at $10^{-4}$.

\begin{table*}[t]
\caption{Seed variation and paired significance tests.
Values are success rate in percent (mean $\pm$ standard deviation over five training seeds).
Confidence intervals and two-sided $p$-values use the hierarchical paired bootstrap described in the text.}
\centering
\resizebox{0.76\linewidth}{!}{
\begin{tabular}{l l c c c c}
\toprule
\textbf{Evaluation} & \textbf{Suite / embodiment} & \textbf{FM} & \textbf{ADS} & \textbf{$\Delta$ 95\% CI} & \textbf{$p$} \\
\midrule
Proposal only & RoboCasa & $54.9\pm0.6$ & $56.4\pm0.7$ & $[0.5,2.5]$ & 0.004 \\
Proposal only & LIBERO-Plus & $76.5\pm0.8$ & $79.4\pm0.7$ & $[1.8,4.1]$ & $<0.001$ \\
Proposal only & WidowX-Bridge & $58.6\pm1.1$ & $62.6\pm1.0$ & $[2.2,5.8]$ & $<0.001$ \\
Proposal only & Google Robot & $69.1\pm0.9$ & $70.1\pm0.8$ & $[0.1,1.9]$ & 0.031 \\
RoboMonkey@$8$ & RoboCasa & $55.5\pm0.7$ & $57.9\pm0.6$ & $[1.1,3.7]$ & $<0.001$ \\
RoboMonkey@$8$ & RoboTwin~2.0 & $89.8\pm0.5$ & $93.5\pm0.4$ & $[2.2,5.1]$ & $<0.001$ \\
\bottomrule
\end{tabular}}
\label{table:supp_statistical_tests}
\end{table*}

\section{Evaluation Metrics and Diagnostic Construction}
\label{app_d}

\paragraph{Proposal spread and training-set replay.}
For $M$ action chunks $a^{(1:M)}\in\mathbb R^{T\times d}$ sampled under one condition, proposal spread is
\begin{equation}
D_{\mathrm{pair}}
=
\sqrt{
\frac{2}{M(M-1)Td}
\sum_{i<j}\|a^{(i)}-a^{(j)}\|_2^2
}.
\end{equation}
Training-set replay uses normalized action coordinates and reports the mean and minimum candidate-to-demonstration distances among $M=8$ samples before averaging over conditions.
For independent candidates at a fixed condition, the unnormalized conditional variance $\mathcal V_\theta(c)=\mathbb E\|a-\mathbb E a\|_2^2$ satisfies
\begin{equation}
\mathbb E[D_{\mathrm{pair}}^2\mid c]=\frac{2\mathcal V_\theta(c)}{Td}.
\label{eq:ads_spread_variance}
\end{equation}
Indeed, independence cancels the cross term in $\mathbb E\|a-a'\|_2^2=2\mathcal V_\theta(c)$, and the pair count cancels the prefactor. For a finite set, the exact identity is $D_{\mathrm{pair}}^2=2\widehat{\mathcal V}/(Td)$, where the unbiased sample variance is $\widehat{\mathcal V}=(M-1)^{-1}\sum_m\|a^{(m)}-\bar a\|_2^2$. Thus pairwise spread and the unnormalized variance are not interchangeable; the normalization and the order of averaging and square rooting must be specified when reporting them.

\paragraph{Geometric redundancy and oracle coverage.}
Let $\rho$ be the mean off-diagonal correlation of centered candidate embeddings. The quantity reported as $M_{\mathrm{eff}}$ is
\begin{equation}
M_{\mathrm{eff}}=
\frac{M}{1+(M-1)\rho}.
\label{eq:supp_meff}
\end{equation}
When $\rho$ is computed from candidate coordinates, this is a geometric redundancy index, not an estimate of the number of statistically independent policy draws. A statistical effective-sample-size interpretation requires covariance across repeated realizations of random variables. Appendix~\ref{app:ads_sampling} states this distinction and derives occupied-mode and oracle-coverage formulas for independent draws. Zero embedding variance makes a correlation undefined, rather than establishing $\rho=1$.
Oracle success at budget $M$ is one when at least one candidate rollout succeeds, and upper-bounds a deployable verifier because outcomes are unavailable during selection.

\paragraph{Entropy gap and mode margin.}
For a sibling set, the normalized entropy gap and the mode margin are
\begin{equation}
\Delta H
=
\frac{H(q)-H(p_{\theta,\cdot\mid0})}{\log K},
\qquad
\Delta_b=Q_b-P_{b\mid0}.
\label{eq:supp_rebalancing}
\end{equation}
Positive $\Delta H$ means the detached target is broader than the primary-conditioned surrogate on that sibling set.

\paragraph{Trace--rollout association.}
Spearman's $\rho$ is computed between $\ell_k$ and binary candidate rollout success within each seed and then averaged across seeds.
The inter-mode invalid rate counts candidates farther than 1.5 pooled within-mode standard deviations from every rollout-validated mode centroid.

\paragraph{Endpoint Jacobian.}
Proposal contraction has a local counterpart: the endpoint Jacobian
\begin{equation}
    J^z_\theta(c,z)
    =
    \frac{\partial\Psi_\theta(z;h_\phi)}{\partial z},
\label{eq:endpoint_jacobian}
\end{equation}
whose shrinking singular values say that distinct base-noise directions no longer produce distinct endpoint directions.
Randomized JVPs estimate the singular values $s_i$ of $J_\theta^z$.
We report $\|J_\theta^z\|_F$, effective rank $(\sum_i s_i^2)^2/\sum_i s_i^4$, and log-volume $\sum_i\log(s_i+10^{-6})$.
These quantities measure endpoint sensitivity to base noise.

\paragraph{Action modes.}
Action chunks are normalized per dimension and embedded by temporal concatenation.
For each fixed condition, $k$-means is run with 20 restarts for $k\in\{2,\ldots,8\}$, and $k$ is chosen by silhouette score; the resulting clusters are the modes used throughout the diagnostics.
Modes with rollout success below 50\% are invalid, and later-checkpoint centroids are matched to the initial centroids with the Hungarian algorithm.
The statistic is population-level occupancy, computed over the sampled set at each checkpoint rather than by tracking individual noise draws.

\begin{table*}[t]
\caption{Sampling protocol for supplementary diagnostics.
Counts are per training seed unless a total is explicitly stated.}
\centering
\resizebox{0.96\linewidth}{!}{
\begin{tabular}{l c c c l}
\toprule
\textbf{Diagnostic} & \textbf{Conditions} & \textbf{Samples/condition} & \textbf{Checkpoints} & \textbf{Aggregation} \\
\midrule
Training-set replay & 2,000 & 8 & final & $d_{\mathrm{mean}}$ and $d_{\mathrm{best}}$ \\
Local $Q-P$ rebalancing & 512 & 8 & $p=0.6,0.7,0.8,0.9,1.0$ & condition mean, then seed mean \\
Trace--rollout validity & 500 & 8 & final & 20,000 pooled candidates; seed bootstrap \\
Endpoint Jacobian & 256 & 4 noise draws, 32 JVPs & final & singular-value statistics per condition \\
Action-mode occupancy & 200 & 64 & five ADS-phase points & centroid matching within condition \\
Candidate-budget scaling & 500 & $M\in\{1,2,4,8,16\}$ & final & paired initial states \\
\bottomrule
\end{tabular}}
\label{table:supp_diagnostic_protocol}
\end{table*}

\section{Training-Compute Accounting}
\label{app:training_compute}

ADS is a training-only procedure and introduces no additional deployment cost: at inference, the proposal policy and downstream verifier are identical to their respective baselines. Its training overhead comes from integrating $K$ sibling trajectories and evaluating $R$ Hutchinson probes for the flow-support estimate. Table~\ref{table:ads_efficiency} reports the resulting training FLOPs, wall-clock step time, peak memory, and deployment FLOPs for different $(K,R)$ configurations. Our default setting, $K=8$ and $R=8$, requires $3.37\times$ the training FLOPs of flow matching, while batching limits the wall-clock overhead to $1.28\times$. A cheaper setting with $K=4$ and $R=4$ reduces the cost to $1.67\times$ training FLOPs while retaining comparable success ($56.1\%$ versus $56.4\%$ for the default configuration), indicating that ADS remains effective under a substantially smaller training budget.

\begin{table}[t]
    \centering
    \caption{Training-compute accounting for ADS.
    Training FLOPs, step time, and peak memory are normalized by standard flow matching under the same hardware and batch size. Deployment FLOPs remain unchanged.}
    \resizebox{0.86\linewidth}{!}{
    \begin{tabular}{l c c c c}
        \toprule
        \textbf{Training method}
        & \textbf{Train FLOPs}
        & \textbf{Step time}
        & \textbf{Peak memory}
        & \textbf{Deployment FLOPs} \\
        \midrule
        Flow matching
        & 1.00$\times$
        & 1.00$\times$
        & 1.00$\times$
        & 1.00$\times$ \\
        ADS ($K=4,R=4$)
        & 1.67$\times$
        & 1.09$\times$
        & 1.15$\times$
        & 1.00$\times$ \\
        ADS ($K=8,R=4$)
        & 2.34$\times$
        & 1.16$\times$
        & 1.25$\times$
        & 1.00$\times$ \\
        \rowcolor{blue!4}
        ADS ($K=8,R=8$, default)
        & 3.37$\times$
        & 1.28$\times$
        & 1.40$\times$
        & 1.00$\times$ \\
        \bottomrule
    \end{tabular}
    }
    \label{table:ads_efficiency}
\end{table}

\section{Discussion and Limitations}

ADS intervenes on the proposal side of test-time verification.
Verifier papers ask how to score candidates; ADS asks whether the candidate set contains alternatives worth scoring.
The empirical pattern matches that framing: the gains appear with RoboMonkey and TokenPB, both of which are built to exploit candidate pools.
Equation~\eqref{eq:supp_cluster_vote} says how: sibling multiplicity determines how much of a sampled set falls in a given mode, and the Hutchinson trace reweights those samples by flow support.

The scope of the claim follows from the same equations.
ADS resamples the variation the current policy expresses. The mode-occupancy measurements support slower disappearance of valid alternatives in the evaluated setting; the objective alone does not guarantee this behavior for every policy. A mode absent from the sibling set receives no target mass in that update, and coincident siblings give zero endpoint correction.
The finite-set decomposition describes the detached target rather than the optimizer's trajectory, and the diagnostics in Appendix~\ref{app_i} measure population-level occupancy rather than a persistent correspondence between a noise region and a semantic mode.

The method has the following limitations.
First, ADS adds training cost: with $K$ siblings, $R$ finite-difference probes, and $S$ trace points, the branch adds roughly $2KRS$ velocity evaluations per update, batched and confined to training.
Second, coincident siblings yield identical weighted moments and a zero endpoint correction, so the method has no effect once endpoint collapse is exact.
Third, the exact base-density-minus-divergence expression is the log density of the continuous flow under the regularity assumptions in Appendix~\ref{app:ads_theory_scope}; its finite-difference and finite-step implementation is used as an approximate relative score and is not a calibrated predictor of task success.
Fourth, pairwise action distance and the geometric redundancy index do not capture semantic equivalence between action chunks, which is why the oracle and mode-coverage analyses accompany them.
Fifth, the Gaussian endpoint gradient is a first-moment correction and can place an update between separated modes; trace weighting reduces the rate of such candidates to 3.1\% (Table~\ref{table:flow_filter_validity}) but the objective contains no explicit repulsion or assignment term.
Sixth, broader proposals hurt when noise or state perturbation makes the local action mode ambiguous, as two LIBERO-Plus splits show.
Finally, ADS depends on routing its gradient to the action expert alone; architectures with shared action-language attention need dedicated adapters or stop-gradient paths.

\section{Training Algorithm}
\label{app_g}
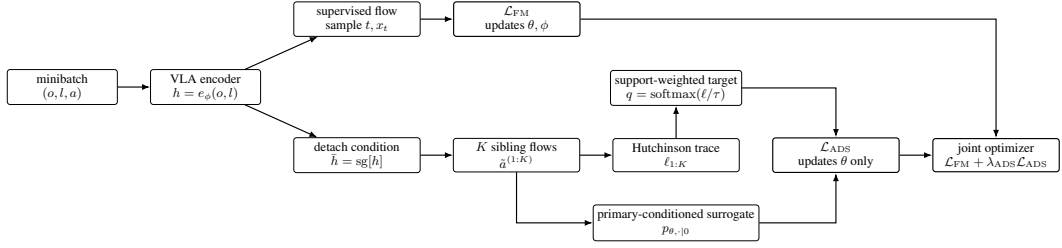
\begin{figure*}[t]
\centering
\small
\resizebox{\textwidth}{!}{%
\begin{tikzpicture}[
    node distance=8mm and 8mm,
    block/.style={rectangle, rounded corners=2pt, draw, align=center, minimum width=27mm, minimum height=8mm},
    branch/.style={rectangle, rounded corners=2pt, draw, align=center, minimum width=31mm, minimum height=8mm},
    loss/.style={rectangle, rounded corners=2pt, draw, thick, align=center, minimum width=31mm, minimum height=8mm},
    arrow/.style={-{Latex[length=2mm]}, thick}
]
\node[block] (batch) {minibatch\\$(o,l,a)$};
\node[block, right=of batch] (enc) {VLA encoder\\$h=e_\phi(o,l)$};
\node[branch, above right=of enc] (fm) {supervised flow\\sample $t,x_t$};
\node[loss, right=of fm] (fmloss) {$\mathcal{L}_{\mathrm{FM}}$\\updates $\theta,\phi$};
\node[branch, below right=of enc] (detach) {detach condition\\$\bar h=\mathrm{sg}[h]$};
\node[branch, right=of detach] (siblings) {$K$ sibling flows\\$\tilde a^{(1:K)}$};
\node[branch, right=of siblings] (trace) {Hutchinson trace\\$\ell_{1:K}$};
\node[branch, above=of trace] (vote) {support-weighted target\\$q=\mathrm{softmax}(\ell/\tau)$};
\node[branch, below=of trace] (surrogate) {primary-conditioned surrogate\\$p_{\theta,\cdot\mid0}$};
\node[loss, right=of trace] (klloss) {$\mathcal{L}_{\mathrm{ADS}}$\\updates $\theta$ only};
\node[loss, right=of klloss] (update) {joint optimizer\\$\mathcal{L}_{\mathrm{FM}}+\lambda_{\mathrm{ADS}}\mathcal{L}_{\mathrm{ADS}}$};
\draw[arrow] (batch) -- (enc);
\draw[arrow] (enc) -- (fm);
\draw[arrow] (fm) -- (fmloss);
\draw[arrow] (enc) -- (detach);
\draw[arrow] (detach) -- (siblings);
\draw[arrow] (siblings) -- (trace);
\draw[arrow] (trace) -- (vote);
\draw[arrow] (siblings) |- (surrogate);
\draw[arrow] (vote) -| (klloss);
\draw[arrow] (surrogate) -| (klloss);
\draw[arrow] (fmloss) -| (update);
\draw[arrow] (klloss) -- (update);
\end{tikzpicture}
}
\caption{ADS training flow. Flow matching anchors task correctness. Once ADS is enabled, detached siblings supply a current-policy reference, and the Hutchinson trace reweights that reference by flow support before the KL update reaches the action expert.}
\label{fig:ds_flow}
\end{figure*}

\begin{algorithm*}[t]
\caption{Flow-Support-Weighted Finite-Set Correction}
\label{alg:ds}
\begin{algorithmic}[1]
\STATE \textbf{Input:} batch $\mathcal{B}$, action expert $v_\theta$, encoder $e_\phi$
\STATE \textbf{Hyperparameters:} $K,R,\tau,\lambda_{\mathrm{ADS}}$, ADS start $p_{\mathrm{ADS}}$, trace start $p_{\mathrm{tr}}$
\STATE \textbf{Output:} updated parameters $\theta,\phi$
\FOR{each minibatch $\mathcal{B}$}
    \STATE Let $p\in[0,1]$ be normalized training progress.
    \STATE Encode task condition $h_i=e_\phi(o_i,l_i)$.
    \STATE Sample flow time $t_i$, base noise $x_{0,i}$, and interpolation state $x_{t,i}$.
    \STATE Compute supervised flow loss $\mathcal{L}_{\mathrm{FM}}$ using Eq.~\eqref{eq:supp_fm}.
    \IF{$p<p_{\mathrm{ADS}}$}
        \STATE Update with $\nabla_{\theta,\phi}\mathcal{L}_{\mathrm{FM}}$ and continue.
    \ENDIF
    \STATE Detach the ADS branch condition $\bar h_i=\mathrm{sg}[h_i]$.
    \STATE Sample primary $z_i^{(0)}$ and integrate $\mu_i=\Psi_\theta(z_i^{(0)};\bar h_i)$.
    \FOR{$k=1,\ldots,K$}
        \STATE Sample $z_i^{(k)}\sim p_{\mathrm{base}}$ and integrate $\tilde a_i^{(k)}=\Psi_\theta(z_i^{(k)};\bar h_i)$.
        \STATE Initialize log support $\ell_i^{(k)}\leftarrow \log p_{\mathrm{base}}(z_i^{(k)})$.
        \IF{$p\ge p_{\mathrm{tr}}$}
            \FOR{each ODE discretization point $s$ on the sibling trajectory}
                \STATE Initialize $\widehat{T}\leftarrow 0$.
                \FOR{$r=1,\ldots,R$}
                    \STATE Draw Hutchinson probe $\epsilon_r$.
                    \STATE $\widehat T\leftarrow\widehat T+\epsilon_r^\top[v_\theta(x_s+\delta\epsilon_r,s,\bar h_i)-v_\theta(x_s-\delta\epsilon_r,s,\bar h_i)]/(2\delta)$.
                \ENDFOR
                \STATE Accumulate $\ell_i^{(k)}\leftarrow \ell_i^{(k)}-(\widehat{T}/R)\Delta s$.
            \ENDFOR
        \ENDIF
    \ENDFOR
    \STATE Form the support-weighted target $q_i=\mathrm{softmax}(\ell_i/\tau)$.
    \STATE Compute primary-conditioned probabilities $p_{\theta,i,\cdot\mid0}$ with Eq.~\eqref{eq:supp_p}.
    \STATE Compute $\mathcal{L}_{\mathrm{ADS}}=\mathrm{KL}(\mathrm{sg}[q_i]\|p_{\theta,i,\cdot\mid0})$.
    \STATE Update with $\nabla_{\theta,\phi}\mathcal{L}_{\mathrm{FM}}+s(p)\lambda_{\mathrm{ADS}}\nabla_{\theta}\mathcal{L}_{\mathrm{ADS}}$.
    \STATE Do not backpropagate $\mathcal{L}_{\mathrm{ADS}}$ into $\phi$.
\ENDFOR
\end{algorithmic}
\end{algorithm*}

Algorithm~\ref{alg:ds} shows both the division of labor and the schedule.
Flow matching trains the VLA policy throughout.
After $p_{\mathrm{ADS}}$, detached siblings supply a resampled current-policy reference; after $p_{\mathrm{tr}}$, the trace term weights that reference by flow support.
The resulting correction reaches the action expert alone.
Figure~\ref{fig:ds_flow} shows the same computation as a data-flow diagram.

\section{Implementation and Hyperparameters}
\label{app_h}
Unless otherwise specified by a benchmark-specific configuration, ADS uses $K=8$ sibling action samples for the KL branch.
The Hutchinson estimator uses $R=8$ probes; the sensitivity analysis additionally evaluates $R\in\{1,2,4\}$.
The centered finite-difference implementation uses radius $\delta=10^{-4}$ in the float16 trace branch.
The flow sampler uses four inference time steps with eight repeated diffusion refinement steps in the GR00T-style configurations.
The KL coefficient $\lambda_{\mathrm{ADS}}$ is implemented as \texttt{alpha\_kl}; the sensitivity configurations use values in $\{10^{-4},3\!\times\!10^{-4},10^{-3},3\!\times\!10^{-3}\}$, with $\lambda_{\mathrm{ADS}}=0$ corresponding to the no-ADS baseline.
The schedule parameters \texttt{lambda\_zero\_ratio} and \texttt{num\_lambdas} control when and how quickly the finite-set term is activated; the default uses $p_{\mathrm{ADS}}=0.5$ and \texttt{num\_lambdas}$=10$.
Trace filtering begins at $p_{\mathrm{tr}}=0.6$; during the short interval $p_{\mathrm{ADS}}\le p<p_{\mathrm{tr}}$, the same detached finite-set KL uses the base-density term alone, without the trace correction.
The temperature $\tau$ in the analysis denotes the softness of the normalized sibling target; it is not swept in the reported fixed-suite evaluations.

The ADS branch is attached after the ordinary flow-matching loss has been constructed.
It samples sibling endpoints from the current policy under $\mathrm{sg}[h_\phi]$, computes support logits $\ell_k(p)$, constructs $q(p)$, and evaluates the finite-set KL against the primary-conditioned surrogate $p_{\theta,\cdot\mid0}$.
All sibling endpoints and $q$ are detached in the KL target.
Consequently, $\mathcal{L}_{\mathrm{ADS}}$ updates the action expert through $\mu_\theta(c,z^{(0)})$ and the local surrogate logits, while $\mathcal{L}_{\mathrm{FM}}$ remains the only loss that updates the VLA encoder.
This is the implementation reason that ADS is better matched to architectures with a separable action expert than to fully shared action-language attention.

\paragraph{Compute accounting.}
Training FLOPs count every batched action-expert evaluation by its tensor dimensions; batching reduces wall time but not arithmetic cost.
The profiler measurements in the main paper use the same hardware, batch size, action horizon, and numerical precision for all rows.
Peak memory includes the encoder activations, sibling trajectories, and finite-difference probes.
Deployment FLOPs are unchanged because the sibling and trace computations are removed after training.
\section{Compute-Matched Training and Entropy Controls}
\label{app:fair_training_controls}

This comparison separates flow-support weighting from additional optimization, early stopping, and entropy-driven proposal spread. We evaluate the controls on RoboCasa, where the component ablation, candidate-budget study, and training-compute measurements use a common policy and task suite. Table~\ref{table:fair_training_controls} reports proposal spread, single-sample success, oracle coverage, and the success retained by each fixed verifier.

\subsection{Shared initialization, data, and model selection}
\label{app:fair_control_protocol}

All methods start from the same pretrained checkpoint for each of the five paired seeds, $0$--$4$, and use the same training and validation splits, observation processing, action normalization, batch size, optimizer, and gradient clipping. The minibatch and augmentation streams share the same seed-specific ordering; additional updates revisit the same training split without adding demonstrations. Separate random-number streams for data, action sampling, and auxiliary probes prevent changes in an auxiliary branch from changing the data order. The action horizon, integration rule, and deployment policy are fixed across methods.

Let $N_0$ denote the standard FM training length and $N_m$ the planned training length of method $m$. Checkpoints are saved at the same five relative update positions, $u/N_m\in\{1/3,1/2,2/3,5/6,1\}$. Hyperparameters are selected by mean validation single-sample success across the paired seeds, and one checkpoint per seed is selected using the same validation criterion. Following Table~\ref{table:supp_optimization}, validation differences below $0.2$ percentage points are broken by lower cumulative training FLOPs. Candidate diversity, oracle outcomes, verifier scores, and test success are not used to select a checkpoint. Each regularized method receives four nonzero weight configurations and the same number of validation evaluations. The equal-FLOPs FM run supplies the shared $\lambda=0$ candidate; selecting it reuses that run. For each entropy control, the weight grid is $\{10^{-4},10^{-3},10^{-2},10^{-1}\}$; uniform-sibling training and ADS use the four KL weights in Table~\ref{table:supp_optimization}. The selected weight is reported with the results. Search compute is recorded separately from the training cost of the selected configuration.

\paragraph{Equal total training FLOPs.}
For update $u$, let $F_{m,u}$ include encoder and action-expert computation, primary and auxiliary integrations, trace evaluation when present, backward passes, and the optimizer update. We accumulate
\begin{equation}
 C_m(n)=C_m^{\mathrm{prep}}+\sum_{u=1}^{n}F_{m,u},\qquad
 B=C_{\mathrm{ADS}}(N_0),\qquad C_0=C_{\mathrm{FM}}(N_0).
\label{eq:fair_total_compute}
\end{equation}
Here $C_m^{\mathrm{prep}}$ accounts for method-specific preparation. Profiling accounts for the activation schedule of each branch, so a per-step overhead is not treated as a constant multiplier for the full training run. Each extended run starts from the common initialization. After the same 500-update warmup, cosine decay spans its full planned training length, rather than extending a run whose learning rate has already decayed to zero. We report the full run cost used for checkpoint selection, even when an earlier checkpoint is selected, together with the selected update $u^*/N_0$. Training wall time is recorded separately on the same hardware.

\paragraph{Early FM checkpoints.}
Table~\ref{table:early_fm_controls} reports all five predefined checkpoints from the standard FM trajectory. These are prefixes of the original schedule, not separately trained models with shortened schedules. The earlier checkpoints directly test whether retaining a less-contracted proposal set also retains executable actions. Their outcomes are reported without choosing a checkpoint by test performance or diversity; the FM row of Table~\ref{table:fair_training_controls} follows the common validation-selection rule above.

\subsection{Entropy objectives and gradient routing}
\label{app:entropy_control_objectives}

The entropy controls minimize
\begin{equation}
 \mathcal L_{\mathrm{FM+Ent}}
 =\mathcal L_{\mathrm{FM}}-s(p)\lambda_{\mathrm E}
 \mathbb E_c[\mathcal E_\theta(c)],
\label{eq:fair_entropy_objective}
\end{equation}
using the same auxiliary-loss activation and ramp $s(p)$ as ADS, with progress measured relative to each run's planned length. The coefficient is selected independently for each objective. All auxiliary branches condition on $\mathrm{sg}[h_\phi]$ and update only the action expert; the ordinary FM term continues to update the same parameters as in ADS. We use $K=8$ and the same flow-integration settings throughout.

\paragraph{Direct entropy of the local surrogate.}
Using the existing local probabilities $p_{\theta,k\mid0}$ from Eq.~(\ref{eq:p}), this baseline maximizes
\begin{equation}
 \mathcal E_{\mathrm{local}}
 =\frac{H(p_{\theta,\cdot\mid0})}{\log K}
 =-\frac{1}{\log K}\sum_{k=1}^{K}
 p_{\theta,k\mid0}\log p_{\theta,k\mid0}.
\label{eq:fair_local_entropy}
\end{equation}
It reuses the fixed covariance, detached sibling endpoints, and differentiable primary endpoint of ADS, and does not evaluate flow support. For $U_k=1/K$,
\begin{equation}
 \mathrm{KL}(p_{\theta,\cdot\mid0}\|U)
 =\log K-H(p_{\theta,\cdot\mid0}),
 \qquad
 \mathrm{KL}(U\|p_{\theta,\cdot\mid0})
 =-\log K-\frac1K\sum_k\log p_{\theta,k\mid0}.
\label{eq:fair_entropy_vs_uniform}
\end{equation}
The uniform-sibling ablation minimizes the second expression. We therefore evaluate it separately from direct local-entropy regularization, while keeping its sibling construction and gradient routing unchanged. Local entropy measures the spread of probability over sampled siblings; the following two controls act directly on the geometry of sampled actions.

\paragraph{Gaussian covariance-entropy proxy.}
Flatten the normalized, pre-discretization action chunks into the rows of $A\in\mathbb R^{K\times D}$, where $D=Td$, and let $A_c=A-\mathbf1\bar a^\top$. With $\widehat C=A_c^\top A_c/(K-1)$, we maximize
\begin{equation}
\begin{aligned}
 \mathcal E_{\mathrm{Cov}}
 &=\frac{1}{2D}\log\det\left(I_D+\frac{\widehat C}{\varepsilon^2}\right)
 =\frac{1}{2D}\log\det\left(I_K+
 \frac{A_cA_c^\top}{(K-1)\varepsilon^2}\right).
\end{aligned}
\label{eq:fair_covariance_entropy}
\end{equation}
The second form follows from the determinant identity and is evaluated with a Cholesky factorization of a $K\times K$ matrix. We fix $\varepsilon=10^{-4}$ in normalized action coordinates. The positive floor makes the expression well defined when $K<D$. This is a Gaussian entropy proxy, up to a fixed additive constant and dimension normalization.

\paragraph{Gaussian-KDE R\'enyi-2 entropy.}
For the same action endpoints, form $\widehat f_h(a)=K^{-1}\sum_k\mathcal N(a;a_k,h^2I_D)$. Its squared-density integral is proportional to $K^{-2}\sum_{i,j}\exp(-\|a_i-a_j\|_2^2/(4h^2))$. At a fixed bandwidth, maximizing its R\'enyi-2 entropy is equivalent to maximizing
\begin{equation}
 \mathcal E_{\mathrm{KDE}}
 =-\frac{1}{\log K}\log\left[
 \frac1{K^2}\sum_{i,j=1}^{K}
 \exp\left(-\frac{\|a_i-a_j\|_2^2}{4h^2}\right)
 \right].
\label{eq:fair_kde_entropy}
\end{equation}
We retain the diagonal terms and evaluate the sum using log-sum-exp. The bandwidth is calibrated once per seed from 512 fixed training conditions at the halfway checkpoint of standard FM: if $m$ is the median non-diagonal squared endpoint distance, then $h=\max\{10^{-4},\sqrt{m/(4\log2)}\}$. This calibration uses no validation or test outcomes and the bandwidth remains fixed during each run. The value of $h$ is recorded with the run configuration. The cost of obtaining the reference FM checkpoint and evaluating these calibration conditions is charged to $C_{\mathrm{KDE}}^{\mathrm{prep}}$, including when the checkpoint is reused from the FM comparison; other methods have zero preparation cost.

Both action-entropy controls retain gradients through all $K$ sampled endpoints and their integration paths. Unlike the primary-only local surrogate, these objectives require differentiable auxiliary trajectories; their backward cost is included in Eq.~(\ref{eq:fair_total_compute}). The covariance matrix and its Cholesky reduction are evaluated in float64; the kernel reduction uses float32. The common backbone precision is unchanged. Neither action-entropy control requires a trace branch. ADS retains the float16 trace implementation and $\delta=10^{-4}$ specified in Appendix~\ref{app_h}.

\subsection{Fixed-verifier evaluation and reporting}
\label{app:fair_control_evaluation}

All rows use the same held-out tasks, paired initial states, evaluation episode counts, action-execution horizon, and continuation protocol. RoboMonkey and TokenPB retain identical checkpoints and scoring settings across proposal methods. Single-sample success uses $M=1$; pairwise spread, oracle coverage, and both verifier results use $M=8$. Oracle and verifier evaluations use the same sampled candidate pool and matched continuation for the comparison in Appendix~\ref{app:ads_sampling}. Results are first aggregated over the benchmark's episode counts and then averaged across the five training seeds, following Appendix~\ref{app:ads_reporting}. Table~\ref{table:fair_training_controls} reports compute, selected hyperparameters, proposal spread, and rollout outcomes jointly. This distinguishes a larger action spread from improved executable coverage and from gains that a fixed verifier can recover. 

\begin{table}[t]
\centering
\caption{\textbf{Training controls on RoboCasa.} Training controls on RoboCasa with measured total training FLOPs. $C/C_0$ is full training cost, $u^*/N_0$ is the mean selected update across seeds, and $\lambda$ is the selected auxiliary weight. Spread is pairwise action RMSE. Success and coverage are percentages averaged over five seeds.}
\label{table:fair_training_controls}
\resizebox{\linewidth}{!}{%
\begin{tabular}{l c c c c c c c c}
\toprule
Method
& $C/C_0$ & $u^*/N_0$ & $\lambda$
& Spread & $S_1$
& Oracle@8 & RoboMonkey@8 & TokenPB@8 \\
\midrule
FM (standard budget)
& 1.00 & 1.00 & --
& 0.003 & 54.9 & 57.6 & 55.5 & 56.4 \\

FM (equal FLOPs)
& 2.37 & 1.67 & --
& 0.002 & 55.9 & 59.9 & 56.9 & 56.8 \\

FM + $H(p)$
& 2.46 & 1.67 & --
& \textbf{0.032} & 42.3 & 53.0 & 50.3 & 51.5 \\

FM + covariance entropy
& 1.49 & 1.67 & --
& 0.027 & 47.1 & 54.4 & 52.2 & 50.5 \\

FM + KDE R\'enyi-2 entropy
& 2.11 & 1.25 & $10^{-4}$
& 0.012 & 45.6 & 58.4 & 52.0 & 52.6 \\
\rowcolor{blue!4}
\textbf{FM + ADS}
& 2.37 & 1.00 & $10^{-3}$
& 0.021 & \textbf{56.4} & \textbf{63.0} & \textbf{57.9} & \textbf{58.9} \\
\bottomrule
\end{tabular}}
\end{table}

\begin{table}[t]
\centering
\caption{Fixed checkpoints along the standard FM trajectory on RoboCasa. All rows retain the original learning-rate schedule. Spread is measured at $M=8$; success and coverage are percentages. No row is selected using its test outcomes or diversity.}
\label{table:early_fm_controls}
\begin{tabular}{cccccc}
\toprule
$u/N_0$ & Spread & $S_1$
& Oracle@8 & RoboMonkey@8 & TokenPB@8 \\
\midrule
$1/3$ & 0.018 & 43.0 & 49.5 & 45.6 & 46.4 \\
$1/2$ & 0.012 & 49.0 & 55.0 & 51.5 & 52.3 \\
$2/3$ & 0.007 & 53.0 & 57.8 & 54.8 & 55.6 \\
$5/6$ & 0.004 & 54.5 & 58.0 & 55.6 & 56.3 \\
$1$   & 0.003 & 54.9 & 57.6 & 55.5 & 56.4 \\
\bottomrule
\end{tabular}
\end{table}

\section{Supplementary Ablations and Diagnostics}
\label{app_i}
\paragraph{Hyperparameter sensitivity.}
Table~\ref{table:ads_hyperparameters} varies one parameter at a time around the default $K=8$, $R=8$, $\tau=1$, $\lambda=10^{-3}$, and trace start $p=0.6$.
The trends saturate for $R=4$--$8$ and show a broad operating region around the default.
Raising $\lambda$ or $\tau$ increases proposal spread and oracle coverage until single-sample success starts to fall, which locates the quality--spread trade-off rather than a monotone gain.

\begin{table*}[t]
    \caption{Sensitivity to the principal ADS hyperparameters on RoboCasa.
    Each block changes one parameter from the default $K=8$, $R=8$, $\tau=1$, $\lambda=10^{-3}$, and trace start $p=0.6$.
    Results show saturation in $R$ and a shallow optimum in the remaining parameters; spread is proposal spread, and entropy is the normalized target entropy $H(q)/\log K$.}
    \centering
    \resizebox{0.93\linewidth}{!}{
    \begin{tabular}{c c c c || c c c c || c c c c}
        \toprule
        \multicolumn{4}{c||}{\textbf{Sibling count $K$}}
        & \multicolumn{4}{c||}{\textbf{Trace probes $R$}}
        & \multicolumn{4}{c}{\textbf{Temperature $\tau$}} \\
        \textbf{$K$} & \textbf{SR} & \textbf{Spread} & \textbf{FLOPs}
        & \textbf{$R$} & \textbf{SR} & \textbf{Spread} & \textbf{FLOPs}
        & \textbf{$\tau$} & \textbf{SR} & \textbf{Spread} & \textbf{Entropy} \\
        \midrule
        2  & 55.7 & .011 & 1.59$\times$
        & 1  & 51.8 & .028 & 1.57$\times$
        & .25 & 55.8 & .015 & 0.42 \\
        4  & 56.1 & .017 & 2.19$\times$
        & 2  & 53.2 & .024 & 1.83$\times$
        & .50 & 56.2 & .019 & 0.58 \\
        \rowcolor{blue!4}
        8  & 56.4 & 0.021 & 3.37$\times$
        & 4  & 56.3 & .023 & 2.34$\times$
        & 1.00 & 56.4 & 0.021 & 0.71 \\
        16 & 56.2 & .024 & 5.74$\times$
        & 8  & 56.4 & 0.021 & 3.37$\times$
        & 2.00 & 56.0 & .024 & 0.84 \\
        \midrule
        \multicolumn{6}{c||}{\textbf{ADS weight $\lambda$}}
        & \multicolumn{6}{c}{\textbf{Trace-start progress $p$}} \\
        \textbf{$\lambda$} & \textbf{SR} & \textbf{Spread} & \textbf{Oracle@$8$} & \multicolumn{2}{c||}{\textbf{Regime}}
        & \textbf{$p$} & \textbf{SR} & \textbf{Spread} & \textbf{Oracle@$8$} & \multicolumn{2}{c}{\textbf{Regime}} \\
        \midrule
        $10^{-4}$ & 55.4 & .008 & 58.7 & \multicolumn{2}{c||}{weak regularization}
        & 0.0 & 55.7 & .025 & 61.6 & \multicolumn{2}{c}{unstable early target} \\
        $3\!\times\!10^{-4}$ & 56.0 & .015 & 61.2 & \multicolumn{2}{c||}{balanced}
        & 0.3 & 56.1 & .023 & 62.5 & \multicolumn{2}{c}{mild early noise} \\
        \rowcolor{blue!4}
        $10^{-3}$ & 56.4 & 0.021 & 63.0 & \multicolumn{2}{c||}{default}
        & 0.6 & 56.4 & 0.021 & 63.0 & \multicolumn{2}{c}{default} \\
        $3\!\times\!10^{-3}$ & 55.6 & .030 & 63.5 & \multicolumn{2}{c||}{largest spread}
        & 0.8 & 56.0 & .014 & 60.5 & \multicolumn{2}{c}{too little trace training} \\
        \bottomrule
    \end{tabular}
    }
    \label{table:ads_hyperparameters}
\end{table*}

\paragraph{Trace score and rollout validity.}
Table~\ref{table:flow_filter_validity} asks whether flow support is associated with candidate rollout success.
Full ADS gives a Spearman correlation of $\rho=0.43$, 2.30 valid modes, and an inter-mode invalid rate of 3.1\%, against 1.72 and 7.6\% for uniform siblings.
The correlation is what justifies using the trace as a relative weight within a sampled set; it is a ranking result rather than a likelihood calibration.

\begin{table}[t]
    \caption{Does trace weighting favor valid rather than arbitrary variation?
    Trace--success correlation is Spearman's $\rho$ between the flow-support score and candidate rollout success.
    The inter-mode invalid rate counts candidates that fall outside every valid action mode.
    Full ADS reaches the highest valid coverage and the lowest rate of unsupported between-mode candidates.}
    \centering
    \resizebox{0.62\linewidth}{!}{
    \begin{tabular}{l c c c}
        \toprule
        \textbf{Proposal strategy}
        & \textbf{Trace--succ. $\rho$}
        & \textbf{Valid modes@$8$}
        & \textbf{Invalid rate $\downarrow$} \\
        \midrule
        Gaussian perturbation & 0.06 & 1.31 & 11.8\% \\
        Uniform siblings & 0.18 & 1.72 & 7.6\% \\
        Base density only & 0.29 & 2.02 & 5.2\% \\
        \rowcolor{blue!4}
        Flow support (full ADS) & \tb{0.43} & \tb{2.30} & \tb{3.1\%} \\
        \bottomrule
    \end{tabular}
    }
    \label{table:flow_filter_validity}
\end{table}

\paragraph{Direction of the local correction.}
Table~\ref{table:local_rebalancing} measures the diagnostics in Eq.~\eqref{eq:supp_rebalancing} after trace activation.
The two entropies separate over training: the primary-conditioned surrogate falls from 0.58 to 0.42 while the support-weighted target holds near 0.71, so mean $\Delta H$ rises from $+0.07$ to $+0.29$ and the fraction of states with $\Delta H>0$ reaches 82.4\%.
This separation measures a difference between two normalized distributions on the same finite reference set. It does not by itself establish invariance of the absolute flow density or an increase in the probability mass of the deployed policy.
At the final checkpoint, 67.3\% of valid non-dominant modes satisfy $Q_b>P_{b\mid0}$. This identifies underweighted modes relative to the local surrogate; Appendix~\ref{app:ads_theory_scope} gives the additional coupling and step-size conditions needed to turn a target discrepancy into a specified surrogate change.

\begin{table*}[t]
    \caption{Local surrogate correction during the ADS phase on RoboCasa.
    Entropies are normalized by $\log K$.
    The final column reports the fraction of valid non-dominant modes for which $Q_b>P_{b\mid0}$.
    Positive entropy gaps and mode margins show that the detached sibling reference is usually broader than the primary-conditioned surrogate.}
    \centering
    \resizebox{0.86\linewidth}{!}{
    \begin{tabular}{c c c c c c}
        \toprule
        \textbf{Training progress $p$}
        & \textbf{$H(p_{\theta,\cdot\mid0})/\log K$}
        & \textbf{$H(q)/\log K$}
        & \textbf{$\mathbb{E}[\Delta H]$}
        & \textbf{$\Pr(\Delta H>0)$}
        & \textbf{Valid minor modes with $Q_b>P_{b\mid0}$} \\
        \midrule
        0.60 & 0.58 & 0.65 & +0.07 & 61.8\% & 54.2\% \\
        0.70 & 0.53 & 0.68 & +0.15 & 69.7\% & 58.6\% \\
        0.80 & 0.47 & 0.71 & +0.24 & 76.8\% & 63.5\% \\
        0.90 & 0.44 & 0.72 & +0.28 & 80.9\% & 66.1\% \\
        \rowcolor{blue!4}
        1.00 & 0.42 & 0.71 & +0.29 & 82.4\% & 67.3\% \\
        \bottomrule
    \end{tabular}
    }
    \label{table:local_rebalancing}
\end{table*}

\paragraph{Noise sensitivity and mode occupancy.}
Table~\ref{table:endpoint_jacobian} reports endpoint sensitivity to base noise.
Full ADS shows the highest Jacobian norm, effective rank, and log-volume, so more latent directions reach the action endpoint.
The uniform-sibling row separates that quantity from usefulness: it attains the largest proposal spread in Table~\ref{table:ads_component_ablation} while scoring worst on rollouts, because sensitivity to noise measures how far candidates spread and not whether the flow supports where they land.
Reading the two together, what distinguishes full ADS is not the amount of endpoint variation but its distribution across latent directions: the highest effective rank despite a smaller proposal spread.
Table~\ref{table:action_cluster_dynamics} measures population-level occupancy of rollout-validated modes under a fixed matching procedure.
The ADS population occupies more valid modes at the final checkpoint; the statistic is computed over the sampled set at each checkpoint rather than by tracking individual noise draws.

\begin{table}[t]
    \caption{Endpoint sensitivity to base noise on RoboCasa.
    We estimate singular values $s_i$ of $J^z_\theta=\partial\Psi_\theta/\partial z$ with randomized JVPs.
    Effective rank is $(\sum_i s_i^2)^2/\sum_i s_i^4$, and log-volume is $\sum_i\log(s_i+10^{-6})$.
    Larger values mean more latent directions reach the action endpoint, which a wider candidate set requires but which does not by itself make those directions executable.}
    \centering
    \resizebox{0.54\linewidth}{!}{
    \begin{tabular}{l c c c}
        \toprule
        \textbf{Training objective}
        & \textbf{$\|J^z_\theta\|_F$}
        & \textbf{Eff. rank}
        & \textbf{Log-vol.} \\
        \midrule
        $\mathcal L_{\mathrm{FM}}$ & 0.18 & 1.18 & $-8.42$ \\
        $+\,$uniform siblings & 0.31 & 1.74 & $-6.37$ \\
        $+\,$base density only & 0.38 & 2.08 & $-5.61$ \\
        \rowcolor{blue!4}
        $+\,$flow support (full ADS) & 0.44 & 2.43 & $-4.92$ \\
        \bottomrule
    \end{tabular}
    }
    \label{table:endpoint_jacobian}
\end{table}

\begin{table}[t]
    \centering
    \caption{Occupancy of valid action modes for fixed training conditions.
    At the start of the analysis phase, normalized action chunks are embedded by temporal concatenation and clustered with $k$-means to define the modes; $k$ is chosen by silhouette score, modes with rollout success below 50\% are marked invalid, and later centroids are matched with the Hungarian algorithm.
    The statistic is population-level occupancy over the sampled set at each checkpoint.}
    \resizebox{0.65\linewidth}{!}{
    \begin{tabular}{c c c c c}
        \toprule
        & \multicolumn{2}{c}{\textbf{Occupied valid modes}}
        & \multicolumn{2}{c}{\textbf{Dominant-mode mass}} \\
        \textbf{ADS-phase progress}
        & \textbf{FM} & \textbf{ADS}
        & \textbf{FM} & \textbf{ADS} \\
        \midrule
        0.00 & 2.21 & 2.21 & 0.52 & 0.52 \\
        0.25 & 1.78 & 2.16 & 0.64 & 0.55 \\
        0.50 & 1.42 & 2.13 & 0.75 & 0.58 \\
        0.75 & 1.18 & 2.11 & 0.85 & 0.61 \\
        \rowcolor{blue!4}
        1.00 & 1.07 & 2.08 & 0.91 & 0.63 \\
        \bottomrule
    \end{tabular}
    }
    \label{table:action_cluster_dynamics}
\end{table}

\section{Independent Sampling, Redundancy, and Candidate Coverage}
\label{app:ads_sampling}

This section distinguishes statistical independence from the coverage of useful actions. All probabilities below condition on a fixed observation and instruction $c$ and a fixed trained parameter vector $\theta$ unless stated otherwise.

\paragraph{Proposition J.1 (independence is preserved by decoding).}
Let $z_1,\ldots,z_M\overset{\mathrm{iid}}{\sim}p_{\mathrm{base}}$ be base-noise draws and let $a_m=\Psi_\theta(z_m;c)$ for a measurable deterministic decoder. Then $a_1,\ldots,a_M$ are conditionally independent and identically distributed. In particular, for any scalar projection $f(a)$ with finite, nonzero variance,
\begin{equation}
\operatorname{Cov}(f(a_i),f(a_j)\mid c,\theta)=0\quad(i\ne j),
\qquad
\operatorname{Var}\!\left(\frac1M\sum_m f(a_m)\mid c,\theta\right)
=\frac{\operatorname{Var}(f(a_1)\mid c,\theta)}{M}.
\label{eq:ads_iid_variance}
\end{equation}
\emph{Proof.} For measurable sets $E_1,\ldots,E_M$, the event $a_m\in E_m$ is the event $z_m\in\Psi_\theta^{-1}(E_m;c)$. Independence of the $z_m$ factors the joint probability into its marginals. The covariance and variance statements follow by expanding the variance of the sum. $\square$

Shrinking the conditional action variance does not change this independence statement. At exact collapse, a correlation coefficient requiring positive variance is undefined. If $c$ is instead randomized and then shared by all candidates, marginal correlations can arise from the shared condition; they do not establish dependence between draws conditional on that condition.

\paragraph{Scope of the correlated-observation identity.}
For scalar random variables $Y_1,\ldots,Y_M$ with common variance $\sigma^2>0$ and common pairwise correlation $\rho$, expansion of the covariance terms gives
\begin{equation}
\operatorname{Var}(\bar Y)
=\frac{\sigma^2}{M}\bigl(1+(M-1)\rho\bigr),
\qquad
M_{\mathrm{eff}}=\frac{M}{1+(M-1)\rho}.
\end{equation}
This is an effective-sample-size identity for estimating a mean from correlated observations. Similarity between two realized action vectors across their coordinates is a different statistic. Applying the same algebraic expression to such similarities yields a geometric redundancy index, not a statistical effective sample size for independent policy sampling. Neither identity alone determines the success of best-of-$M$ selection.

\paragraph{Proposition J.2 (expected occupied modes).}
Let $\mathcal C_1,\ldots,\mathcal C_B$ be disjoint, measurable action regions fixed independently of the evaluated candidate set, and write $\alpha_b=\Pr(a\in\mathcal C_b\mid c)$. Let $\mathcal B_{\mathrm{valid}}$ index regions satisfying a separately defined validity criterion. The number $N_M$ of valid regions reached by $M$ independent draws satisfies
\begin{align}
\mathbb E[N_M\mid c]
&=\sum_{b\in\mathcal B_{\mathrm{valid}}}\bigl[1-(1-\alpha_b)^M\bigr],
\label{eq:ads_occupied_modes}\\
\mathbb E[N_{M+1}-N_M\mid c]
&=\sum_{b\in\mathcal B_{\mathrm{valid}}}\alpha_b(1-\alpha_b)^M.
\label{eq:ads_coverage_increment}
\end{align}
\emph{Proof.} Define $I_b=\mathbf 1\{\exists m:a_m\in\mathcal C_b\}$. Independence gives $\Pr(I_b=0\mid c)=(1-\alpha_b)^M$. Since $N_M=\sum_b I_b$, linearity of expectation proves the first identity without assuming that the $I_b$ are independent. Subtracting the identities at $M+1$ and $M$ proves the second. $\square$

For a fixed total mass on a fixed number of valid regions, the expected coverage is a symmetric concave function of the regional masses: for $M\ge2$, the second derivative of $1-(1-\alpha)^M$ is $-M(M-1)(1-\alpha)^{M-2}\le0$. Thus redistributing mass from a more frequent valid region to a less frequent one, without reversing their order or losing total valid mass, increases expected coverage. This is a conditional statement about distributions; it does not assert that an arbitrary ADS parameter update performs that redistribution.

\paragraph{Proposition J.3 (oracle success and conditional heterogeneity).}
Fix an execution rule for the initial candidate chunk and a continuation controller. Suppose candidate rollouts, including any continuation randomness, are independent conditional on $c$, and let their common success probability be $s(c)$. An oracle that selects a successful member of this same evaluated pool has
\begin{equation}
\operatorname{Oracle}_M(c)=1-(1-s(c))^M.
\label{eq:ads_oracle_iid}
\end{equation}
For a distribution of conditions,
\begin{equation}
\mathbb E_c[\operatorname{Oracle}_M(c)]
=\mathbb E_c[1-(1-s(c))^M]
\le 1-(1-\mathbb E_c[s(c)])^M.
\label{eq:ads_oracle_jensen}
\end{equation}
\emph{Proof.} All $M$ independent rollouts fail with probability $(1-s(c))^M$. For $M\ge2$, the function $1-(1-s)^M$ is concave, so Jensen's inequality gives the last inequality; $M=1$ gives equality. $\square$

The realized maximum of binary outcomes upper-bounds the outcome of any selector choosing from those same evaluated outcomes, even without independence. Independence is needed for Eq.~\eqref{eq:ads_oracle_iid}, not for this pointwise maximum property. Reusing a common continuation random seed, changing the continuation controller between methods, or selecting an oracle action at every future step defines a different experiment and must be specified separately. Slow oracle scaling can reflect low mass on useful alternatives, heterogeneous condition difficulty, or execution stochasticity; it does not by itself identify proposal contraction as the unique cause.

\section{Finite-Set Multiplicity and Flow-Support Weighting}

We formalize the decomposition of the finite sibling target into sample multiplicity and flow-support weighting.
Suppose the $K$ sibling actions can be partitioned into $B$ modes
\begin{equation}
    \mathcal{A}_K=\bigcup_{b=1}^{B}\mathcal{C}_b,
    \qquad
    n_b=|\mathcal{C}_b|.
\end{equation}
Each mode groups locally coherent action chunks, such as similar grasp approaches or waypoint sequences; the partition is an analysis device and is not used by ADS.
Define the ADS target mass assigned to mode $b$ as
\begin{equation}
    Q_b=\sum_{k\in\mathcal{C}_b}q_k.
\end{equation}
Using Eq.~\eqref{eq:supp_q}, we have
\begin{equation}
    Q_b
    =
    \frac{
    \sum_{k\in\mathcal{C}_b}\exp(\ell_k/\tau)
    }{
    \sum_{j=1}^{B}\sum_{i\in\mathcal{C}_j}\exp(\ell_i/\tau)
    }.
\label{eq:cluster_vote_exact}
\end{equation}
If the trace-based log-support values are approximately constant inside each mode, $\ell_k=\bar{\ell}_b+\eta_k$ with $|\eta_k|\le\epsilon_b$, then
\begin{equation}
    Q_b
    \approx
    \frac{
    n_b\exp(\bar{\ell}_b/\tau)
    }{
    \sum_{j=1}^{B} n_j\exp(\bar{\ell}_j/\tau)
    }.
\label{eq:supp_cluster_vote}
\end{equation}
More precisely, the numerator of Eq.~\eqref{eq:cluster_vote_exact} lies between
\begin{equation}
    n_b\exp((\bar{\ell}_b-\epsilon_b)/\tau)
    \quad\text{and}\quad
    n_b\exp((\bar{\ell}_b+\epsilon_b)/\tau),
\end{equation}
so the approximation carries a multiplicative error controlled by $\exp(\pm\epsilon_b/\tau)$ before normalization.
Equation~\eqref{eq:supp_cluster_vote} splits the target mass into two factors.
The count $n_b$ is the sampling term: a mode represented by more siblings receives more mass in that draw.
The factor $\exp(\bar{\ell}_b/\tau)$ is the flow-density weighting term. Its usefulness as a task-quality signal depends on the measured association with rollout validity; a large density value alone is not a certificate of executability.
Trace weighting thus redistributes the variation present in the sampled set rather than generating new variation.

This is also what separates ADS from direct action averaging.
Averaging computes a single command
\begin{equation}
    \bar a_{\mathrm{avg}}=\frac{1}{K}\sum_{k=1}^{K}\tilde a^{(k)},
\end{equation}
which can fall between modes and become invalid.
ADS instead forms finite-set target masses $Q_b$ and trains the local surrogate to reduce its discrepancy from that target. The inequality $Q_b>P_b$ gives the sign of an independently parameterized mode-logit derivative; it is not sufficient on its own to establish that the implemented endpoint update raises $P_b$ or the actual policy mass.
The finite-set target changes the training signal and is never executed as an action.
The endpoint gradient remains a first-moment correction, however, so an update can still land between separated modes; Table~\ref{table:flow_filter_validity} measures how often the resulting candidates fall outside every valid mode.

\section{KL Gradients and the Scope of Relative Rebalancing}

For readability, we suppress the primary-trajectory index $\mid0$ in this derivation.
Let $p_{\theta,k}=\exp(r_{\theta,k})/\sum_j\exp(r_{\theta,j})$ and
\begin{equation}
\mathcal{L}_{\mathrm{ADS}}
=
-\sum_{k=1}^{K}q_k\log p_{\theta,k}
+\mathrm{const}(q).
\end{equation}
The softmax derivative gives
\begin{equation}
\frac{\partial \log p_{\theta,i}}{\partial r_{\theta,k}}
=
\mathbf{1}[i=k]-p_{\theta,k}.
\end{equation}
Therefore
\begin{align}
\frac{\partial \mathcal{L}_{\mathrm{ADS}}}{\partial r_{\theta,k}}
&=
-\sum_i q_i(\mathbf{1}[i=k]-p_{\theta,k}) \\
&=
p_{\theta,k}-q_k.
\end{align}
If the logits were independent optimization variables, gradient descent would increase each logit for which $q_k>p_{\theta,k}$. Here $p_{\theta,k}$ is a local surrogate weight, and the implemented logits are coupled through the primary endpoint.
Aggregating by mode gives the same algebraic result under a mode-logit abstraction.
Let
\begin{equation}
    P_b=\sum_{k\in\mathcal{C}_b}p_{\theta,k}.
\end{equation}
If we parameterize a mode logit $R_b$ whose softmax probability is $P_b$, the cross-entropy part of Eq.~\eqref{eq:supp_kl} satisfies
\begin{equation}
    \frac{\partial \mathcal{L}_{\mathrm{ADS}}}{\partial R_b}
    =
    P_b-Q_b.
\label{eq:mode_gradient}
\end{equation}
Under this abstraction, $Q_b>P_b$ specifies the sign of the derivative with respect to $R_b$. Because probabilities are normalized jointly, increasing a single logit and increasing its normalized mass under a simultaneous update are distinct statements. The implemented gradient has the stronger endpoint coupling analyzed in Appendix~\ref{app:ads_theory_scope}.
The implementation carries no explicit mode logits; its actual update is the endpoint moment correction below.

For intuition, consider an isotropic Gaussian surrogate with fixed variance $\sigma^2$ and logits
\begin{equation}
    r_{\theta,k}=-\frac{1}{2\sigma^2}\|\tilde a^{(k)}-\mu_\theta\|_2^2.
\end{equation}
Then
\begin{align}
    \frac{\partial \mathcal{L}_{\mathrm{ADS}}}{\partial \mu_\theta}
    &=
    \frac{1}{\sigma^2}
    \left(
    \sum_{k=1}^{K}p_{\theta,k}\tilde a^{(k)}
    -
    \sum_{k=1}^{K}q_k\tilde a^{(k)}
    \right).
\end{align}
The ADS update corrects the local proposal center by the difference between the surrogate-weighted and support-weighted sibling moments.
In a flow-matching policy this update applies to the denoising dynamics rather than to a final averaged command.
If all sibling endpoints coincide at $a^\star$, both moments equal $a^\star$ and the endpoint gradient vanishes, which is the fixed point that requires ADS to act before contraction completes.

\section{Scope and Sufficient Conditions for the ADS Correction}
\label{app:ads_theory_scope}

This section separates three objects: the density of the generated action,
the probability mass of an action region, and the primary-conditioned
distribution on a sampled reference set. It then gives sufficient conditions
under which an ADS step decreases its detached objective, increases a
specified surrogate mass, and preserves endpoints inside a valid region.
These are local, conditional statements; the validity of an action region
must be established independently of its flow density.

\subsection{Density, mode mass, and the finite-set reference}

Fix a condition $c$ and a policy parameter $\theta_0$. Write
$\pi(a)=\pi_{\theta_0}(a\mid c)$ for the generated density on the continuous
action space $\mathbb R^D$, where $D=Td$. Let $C_b$ be a fixed measurable
action region and let $m_b=\int_{C_b}\pi(a)\,da$ be its population mass.
Independently draw siblings $a_k\sim\pi$, and define
$n_b=\sum_k\mathbf 1[a_k\in C_b]$ and
$Q_b=\sum_{k:a_k\in C_b}q_k$. In contrast,
$P_b=\sum_{k:a_k\in C_b}p_k$ denotes a mass under the Gaussian
\emph{finite-set surrogate}. In general, neither $P_b$ nor $Q_b$ equals
$m_b$. Throughout an update, the sibling locations, their region membership,
and the target weights are held fixed.

\paragraph{Proposition 1 (interpretation of the exact trace score).}
Assume the velocity field is continuously differentiable in its state,
its ODE has a unique solution on $[0,1]$, and the resulting map
$\Psi_{\theta_0}$ is a differentiable bijection with nonsingular Jacobian.
For a base density $p_{\mathrm{base}}$ and an exact trajectory
$x_t$ from $z$ to $a$, the fully activated score satisfies
\begin{equation}
\ell(a)=\log p_{\mathrm{base}}(z)
-\int_0^1\operatorname{Tr}\!\left(\partial_xv_{\theta_0}(x_t,t,c)\right)dt
=\log\pi(a).
\label{eq:ads_exact_density}
\end{equation}
\paragraph{Proof.}
The variational equation for $A_t=\partial x_t/\partial z$ is
$\dot A_t=(\partial_xv_{\theta_0})A_t$, with $A_0=I$.
Jacobi's formula gives
$\frac{d}{dt}\log|\det A_t|=\operatorname{Tr}(\partial_xv_{\theta_0})$.
Substitution into the change-of-variables formula proves the claim.
This is the standard continuous-normalizing-flow identity
\citep{chen2018node,grathwohl2019ffjord}.

Thus, ``flow support'' denotes a density-derived score, rather than a
separate mathematical support set or a certificate of rollout validity.
Low mode mass can coexist with high pointwise density when the mode has
small volume; such coexistence is not an invariance of density under
fine-tuning. Equation~\eqref{eq:ads_exact_density} concerns continuous
endpoints before noninvertible clipping, rounding, or gripper
postprocessing. Finite differences, stochastic trace estimates, and ODE
discretization introduce estimation error. During the base-density-only
phase, the endpoint-density interpretation does not apply.

\paragraph{Proposition 2 (population limit of the weighted reference).}
Let $\tau>0$ and suppose
$0<Z_\tau=\int\pi(a)^{1+1/\tau}\,da<\infty$. For exact scores and every
bounded measurable $f$, the self-normalized reference satisfies, almost surely,
\begin{equation}
\sum_{k=1}^Kq_k f(a_k)
\ \longrightarrow\
\frac{\int f(a)\pi(a)^{1+1/\tau}\,da}{Z_\tau}.
\label{eq:ads_population_target}
\end{equation}
\paragraph{Proof.}
The numerator divided by $K$ is an empirical mean of
$f(a_k)\pi(a_k)^{1/\tau}$, whose absolute expectation is finite.
The denominator divided by $K$ converges to the positive number $Z_\tau$.
The strong law of large numbers and division yield the result.

The reference therefore emphasizes high-density actions already generated
by the policy. For example, for a density constant on disjoint regions of
volumes $v_b$ and masses $m_b$, its limiting region weights are proportional
to $m_b^{1+1/\tau}v_b^{-1/\tau}$. Equal-volume minority regions are not
automatically upweighted. The useful comparison in an ADS step is between
this sampled reference and the \emph{primary-conditioned} surrogate;
Equation~\eqref{eq:ads_population_target} is not the distribution of the
policy after optimization.

\paragraph{Proposition 3 (visibility and finite-set mass bounds).}
For independent siblings,
\begin{equation}
\Pr(n_b>0)=1-(1-m_b)^K,
\qquad
\mathbb E[n_b]=Km_b.
\label{eq:ads_mode_visibility}
\end{equation}
In particular, $K\ge \log(1/\alpha)/m_b$ is sufficient for visibility
probability at least $1-\alpha$ when $m_b>0$ and $0<\alpha<1$.
Suppose the sampled reference is partitioned into regions and its scores
satisfy $|\ell_k-\bar\ell_b|\le\epsilon_b$ within region $b$.
Set $A_b=n_b e^{\bar\ell_b/\tau}$,
$L_b=A_b e^{-\epsilon_b/\tau}$, and
$U_b=A_b e^{\epsilon_b/\tau}$. Then
\begin{equation}
\frac{L_b}{L_b+\sum_{j\ne b}U_j}
\ \le Q_b\le\
\frac{U_b}{U_b+\sum_{j\ne b}L_j}.
\label{eq:ads_normalized_mode_bounds}
\end{equation}
For $\epsilon=\max_b\epsilon_b$ and
$\bar Q_b=A_b/\sum_j A_j>0$, this implies
$e^{-2\epsilon/\tau}\le Q_b/\bar Q_b\le e^{2\epsilon/\tau}$.
\paragraph{Proof.}
The count $n_b$ is binomial, and
$(1-m_b)^K\le e^{-Km_b}$. Each region's unnormalized weight lies in
$[L_b,U_b]$. A normalized mass increases with its own weight and decreases
with every other weight, giving the stated bounds. Applying the common
error bound to both numerator and denominator gives the last inequality.

For $0<P_b<1$, a directly checkable sufficient condition for $Q_b>P_b$ is
\begin{equation}
L_b>\frac{P_b}{1-P_b}\sum_{j\ne b}U_j.
\label{eq:ads_target_surplus}
\end{equation}
This identifies a target surplus despite limited multiplicity. It requires
the region to be sampled, and is a statement about the detached target,
not yet about the update direction. The same error argument applies to
numerically estimated scores: if
$\max_k|\widehat\ell_k-\ell_k|\le\varepsilon$, then
$e^{-2\varepsilon/\tau}\le\widehat q_k/q_k\le e^{2\varepsilon/\tau}$.
Smaller temperatures therefore require more accurate scores. A trace
estimator's unbiasedness alone does not establish this error bound.

\subsection{What an endpoint-coupled update can guarantee}

Throughout this subsection, $\Sigma$ is fixed, symmetric, and positive definite.
For the remainder of this subsection, write
$F(\theta)=\mathrm{KL}(q\|p_\theta)$ for one frozen reference set.
Let $r(\theta)\in\mathbb R^K$ be its surrogate logits,
$J=\partial r/\partial\theta$, $G=JJ^\top$, and $\delta=q-p$,
all evaluated before the update. Then
\begin{equation}
g:=\nabla_\theta F=-J^\top\delta,
\qquad
\theta^+=\theta-\eta g,
\qquad
r(\theta^+)-r(\theta)=\eta G\delta+O(\eta^2).
\label{eq:ads_coupled_step}
\end{equation}
The matrix $G$ exposes the coupling between logits. For the Gaussian
surrogate, the $k$th row of $J$ is
$(a_k-\mu_\theta)^\top\Sigma^{-1}\partial_\theta\mu_\theta$;
there are no independently optimized sibling logits.

\paragraph{Proposition 4 (directional surrogate-mass condition).}
For a nonempty sampled region, define the vector
$(w_b)_k=\mathbf 1[a_k\in C_b]p_k/P_b$. If the logits are twice
continuously differentiable, the first-order changes are
\begin{align}
\log P_b(\theta^+)-\log P_b(\theta)
&=\eta\nu_b+O(\eta^2),
&\nu_b&=(w_b-p)^\top G\delta,
\label{eq:ads_surrogate_mass_direction}\\
\log\frac{P_b(\theta^+)}{P_d(\theta^+)}-
\log\frac{P_b(\theta)}{P_d(\theta)}
&=\eta\nu_{bd}+O(\eta^2),
&\nu_{bd}&=(w_b-w_d)^\top G\delta.
\label{eq:ads_surrogate_odds_direction}
\end{align}
Thus $\nu_b>0$, rather than $Q_b>P_b$ alone, is sufficient for
surrogate mass to increase under a sufficiently small gradient step.
\paragraph{Proof.}
Differentiating the softmax gives
$\partial_r\log P_b=w_b-p$. The chain rule and
$-g=J^\top\delta$ give Equation~\eqref{eq:ads_surrogate_mass_direction}.
Subtracting two log-mass derivatives gives
Equation~\eqref{eq:ads_surrogate_odds_direction}.

These conditions can be tested using gradient inner products without
constructing $G$: $\nu_b=-\langle\nabla_\theta\log P_b,g\rangle$.
They distinguish a positive target surplus from an accessible correction
direction. A nonzero surplus can have $\nu_b\le0$ because of parameter
sharing or because other siblings demand incompatible movements.

\paragraph{Proposition 5 (local descent and preservation of valid endpoints).}
Assume $g\ne0$. Along a candidate parameter segment, let $\nabla F$ be
$L$-Lipschitz and
$\|\nabla^2\log P_b\|_{\mathrm{op}}\le H_b$. Consider a finite set
$\mathcal Z$ of audited base-noise draws. Each endpoint
$\Psi_\theta(z)$ lies in an independently specified valid region $V_z$
with margin
$m_z=\operatorname{dist}(\Psi_\theta(z),V_z^c)>0$.
Suppose
$\|\partial_\theta\Psi_{\theta'}(z)\|_{\mathrm{op}}\le B_z$
along this segment. If $\nu_b>0$, every sufficiently small step satisfying
\begin{equation}
0<\eta<\min\left\{
\frac{2}{L},\;
\frac{2\nu_b}{H_b\|g\|_2^2},\;
\min_{z\in\mathcal Z}\frac{m_z}{B_z\|g\|_2}
\right\}
\label{eq:ads_joint_step_condition}
\end{equation}
and remaining in that segment strictly decreases $F$, increases $P_b$,
and keeps every audited endpoint inside its corresponding $V_z$.
Terms with zero denominator and positive numerator impose no restriction.
\paragraph{Proof.}
The descent lemma and Taylor's theorem respectively give
\begin{align}
F(\theta^+)-F(\theta)
&\le-\eta\left(1-\frac{L\eta}{2}\right)\|g\|_2^2,
\label{eq:ads_frozen_kl_descent}\\
\log P_b(\theta^+)-\log P_b(\theta)
&\ge\eta\nu_b-\frac{H_b\eta^2}{2}\|g\|_2^2.
\label{eq:ads_finite_mass_gain}
\end{align}
The fundamental theorem of calculus yields
$\|\Psi_{\theta^+}(z)-\Psi_\theta(z)\|_2
\le\eta B_z\|g\|_2<m_z$. The ball of radius $m_z$ about the original
endpoint is contained in $V_z$, proving the final statement.

This proposition provides simultaneous, positive local guarantees when
the correction direction and a validity margin are available. It does
not infer validity from density or from a centroid-distance threshold.
It applies to the audited endpoints; a population guarantee requires
the same bounds on a set of base draws with known probability. For
example, if they hold uniformly on $E$ with
$p_{\mathrm{base}}(E)\ge1-\beta$, then the updated probability of lying
in the corresponding valid regions is at least $1-\beta$.

\paragraph{An explicit endpoint smoothness bound.}
If $\mu$ itself is optimized and $\Sigma$ is fixed, put
$b_k=\Sigma^{-1}a_k$. After cancelling a common logit term,
$r_k=\mu^\top b_k-\frac12a_k^\top\Sigma^{-1}a_k$. Consequently,
\begin{equation}
\nabla_\mu^2F=\operatorname{Cov}_{p}(b_k),
\qquad
\|\nabla_\mu^2F\|_{\mathrm{op}}
\le\frac14\max_{i,j}\|b_i-b_j\|_2^2.
\label{eq:ads_endpoint_curvature}
\end{equation}
For any unit direction, the scalar projection of $b_k$ has range at most
the displayed diameter and variance at most one quarter of its squared
range. This proves the bound and convexity in $\mu$. It gives a computable
step-size bound for an endpoint diagnostic, although composing with a
neural parameterization need not preserve convexity.

\paragraph{Joint training and optimizer directions.}
The reference is refreshed after each step, so descent of frozen $F$
does not imply monotonic descent for the rebuilt targets. For an arbitrary
actual parameter direction $d$, including one produced by a joint loss
or an adaptive optimizer, the sufficient descent condition becomes
\begin{equation}
\langle g,d\rangle<0,
\qquad
0<\eta<\frac{-2\langle g,d\rangle}{L\|d\|_2^2}.
\label{eq:ads_general_direction}
\end{equation}
The surrogate-mass condition is
$\langle\nabla\log P_b,d\rangle>0$, with the corresponding Taylor
bound, and the validity displacement bound uses $\eta B_z\|d\|_2$.
For ordinary gradient descent on
$\mathcal L_{\mathrm{FM}}+\lambda F$, the first inner product is
$-\langle g,\nabla\mathcal L_{\mathrm{FM}}\rangle-\lambda\|g\|_2^2$.
Thus the supervised gradient, momentum, and weight decay must be included
when testing these conditions for an implemented training step.

\subsection{Why surrogate improvement is not a mode-mass theorem}

\paragraph{A two-region example.}
Consider one-dimensional frozen siblings $\{-a,+a\}$, $a>0$, equal
target weights, and $\Sigma=\sigma^2$. Direct differentiation gives
\begin{equation}
p_+(\mu)=\frac{1}{1+e^{-2a\mu/\sigma^2}},
\qquad
\frac{dF}{d\mu}=\frac{a}{\sigma^2}
\tanh\!\left(\frac{a\mu}{\sigma^2}\right).
\label{eq:ads_two_region_example}
\end{equation}
The unique minimizer is $\mu=0$, where the surrogate matches the target.
For $\mu\ne0$, the negative gradient points toward this midpoint.
If valid actions occupy disjoint neighborhoods of $-a$ and $+a$, the
minimizer is invalid. In particular, the finite-set target may have higher
entropy while the endpoint correction reduces endpoint separation.
The margin condition in Proposition~5 prevents an individual small step
from leaving its valid neighborhood; it does not permit arbitrarily many
such steps without rechecking the margin. Unequal support can change the
destination but cannot replace that check.

\paragraph{Proposition 6 (additional condition for actual mode-mass gain).}
Let $\theta(t)=\theta+td$ induce differentiable densities $\pi_t$ and
diffeomorphic endpoint maps. Fix a bounded region $C$ with continuously
differentiable boundary and outward unit normal $n$. Assume the densities
and the induced action velocity
\begin{equation}
u(a)=\left.\partial_\theta\Psi_\theta(z)d
\right|_{z=\Psi_\theta^{-1}(a)}
\label{eq:ads_parameter_transport}
\end{equation}
are sufficiently regular to differentiate under the integral and apply
the divergence theorem. Then the actual population mass satisfies
\begin{equation}
\left.\frac{d}{dt}\pi_t(C)\right|_{t=0}
=-\int_{\partial C}\pi_0(a)u(a)^\top n(a)\,dS(a).
\label{eq:ads_population_flux}
\end{equation}
Strictly positive net inward flux is sufficient for an actual mass
increase for all sufficiently small positive $t$.
\paragraph{Proof.}
Transport of the fixed base measure by the parameterized endpoint map
gives the continuity equation
$\partial_t\pi_t=-\nabla\cdot(\pi_tu_t)$.
Integration over $C$ and the divergence theorem give
Equation~\eqref{eq:ads_population_flux}. Differentiability and a strictly
positive derivative imply the final statement.

This condition concerns action-space motion near the boundary of a fixed
region. A surplus $Q_b-P_b$, a positive surrogate derivative $\nu_b$,
or a decrease of $F$ does not determine this boundary flux. Establishing
population mode retention therefore requires either additional assumptions
linking those quantities to the induced transport or direct measurements
of occupancy under fixed region definitions. The preceding results instead
justify a practical sequence of checks: visibility of the relevant region,
reliable within-set weighting, an accessible correction direction, and a
step compatible with independently assessed validity margins.

\subsection{Numerical stability of trace scores and finite-set weights}
\label{app:trace_stability}

Fix a flow time and conditioning input, and write $v:\mathbb R^D\to\mathbb R^D$ and $J=Dv(x)$.
For a probe $\epsilon$, define the scalar projected central difference
\begin{equation}
T_\delta(x,\epsilon)
=\epsilon^\top\frac{v(x+\delta\epsilon)-v(x-\delta\epsilon)}{2\delta}.
\end{equation}
The following bounds distinguish stochastic trace estimation from finite-difference and integration error.

\paragraph{Proposition (trace approximation and weight stability).}
Suppose $v$ is $C^3$ on the segment $\{x+s\epsilon:|s|\le\delta\}$ and its third derivative satisfies
$\|D^3v(y)[u_1,u_2,u_3]\|_2\le B\prod_{j=1}^3\|u_j\|_2$ there.
Then
\begin{equation}
\left|T_\delta(x,\epsilon)-\epsilon^\top J\epsilon\right|
\le \frac{B\delta^2}{6}\|\epsilon\|_2^4.
\label{eq:trace_fd_error}
\end{equation}
This is a bound on the scalar projection, rather than on the unprojected vector difference.
For $R$ independent Rademacher probes with independent coordinates, let
$\widehat T=R^{-1}\sum_{r=1}^R\epsilon_r^\top J\epsilon_r$ and $A=(J+J^\top)/2$.
The exact quadratic estimator satisfies
\begin{equation}
\mathbb E[\widehat T]=\operatorname{tr}J,
\qquad
\operatorname{Var}(\widehat T)
=\frac{2}{R}\sum_{i\ne j}A_{ij}^2.
\label{eq:trace_exact_variance}
\end{equation}
If the derivative bound holds for every sampled probe segment, its finite-difference replacement obeys
$|R^{-1}\sum_r T_\delta(x,\epsilon_r)-\widehat T|\le B\delta^2D^2/6$.
Consequently, the finite-difference estimator need not be unbiased even though the exact quadratic estimator is unbiased.

For any finite score vectors $\ell,\widehat\ell$ with
$\|\widehat\ell-\ell\|_\infty\le e$, define
$q=\operatorname{softmax}(\ell/\tau)$ and
$\widehat q=\operatorname{softmax}(\widehat\ell/\tau)$, where $\tau>0$.
Then, for every sibling $k$,
\begin{equation}
e^{-2e/\tau}\le\frac{\widehat q_k}{q_k}\le e^{2e/\tau},
\qquad
\mathrm{KL}(q\|\widehat q)\le\frac{2e}{\tau}.
\label{eq:trace_weight_stability}
\end{equation}
Moreover, $\ell_i-\ell_j>2e$ implies $\widehat\ell_i>\widehat\ell_j$.

\paragraph{Proof.}
Set $g(s)=\epsilon^\top v(x+s\epsilon)$.
The assumed operator bound gives $|g'''(s)|\le B\|\epsilon\|_2^4$.
Taylor's theorem applied at $\delta$ and $-\delta$, followed by division by $2\delta$, proves Eq.~\eqref{eq:trace_fd_error}.
Since $\epsilon^\top J\epsilon=\epsilon^\top A\epsilon$ and Rademacher coordinates have unit squares,
$\epsilon^\top A\epsilon=\operatorname{tr}A+2\sum_{i<j}A_{ij}\epsilon_i\epsilon_j$.
Distinct unordered pairs have zero covariance, giving variance $4\sum_{i<j}A_{ij}^2$ for one probe; independence across probes proves Eq.~\eqref{eq:trace_exact_variance}.
Also $\|\epsilon\|_2^4=D^2$.
Writing $\eta_k=\widehat\ell_k-\ell_k$ yields
\[
\frac{\widehat q_k}{q_k}
=\frac{e^{\eta_k/\tau}}{\sum_jq_j e^{\eta_j/\tau}}.
\]
Both numerator and denominator lie in $[e^{-e/\tau},e^{e/\tau}]$.
The ratio bound follows, and averaging $\log(q_k/\widehat q_k)\le2e/\tau$ under $q$ proves the KL bound.
Finally, $(\widehat\ell_i-\widehat\ell_j)-(\ell_i-\ell_j)=\eta_i-\eta_j\ge-2e$ proves the ranking statement.

For integrated flow scores, $e$ must account jointly for trajectory discretization, time quadrature, finite differences, stochastic probes, and floating-point arithmetic; the pointwise variance formula alone does not bound that combined error.
Finite-step integration and the nonlinear softmax do not inherit an unbiasedness guarantee.
The implementation uses the float16 trace branch and $\delta=10^{-4}$ specified in Appendix~\ref{app_h}. Reducing $\delta$ decreases the truncation bound but can increase floating-point cancellation error. These bounds characterize sources of numerical error; they do not establish score or ranking stability on the trained network.

\section{Controlled Synthetic Diagnostics}
\label{app:ads_synthetic}

\paragraph{Scope and reproducibility.}
We evaluate three controlled synthetic diagnostics to separate statistical
independence, geometric coverage, density weighting, and endpoint optimization.
These diagnostics use known scalar distributions or fixed surrogate objectives;
they do not train or evaluate a VLA, a flow network, a simulator, or a robot.
Consequently, their outcomes cannot establish improvements in manipulation
success or validate the proposed mechanism in a learned action model.
The accompanying NumPy script uses float64 arithmetic, root seed
$20260923$, and $100{,}000$ independently generated candidate sets per
Monte Carlo configuration. Separate child random streams are used for the
first two diagnostics. The script exports all diagnostic summaries and
optimization trajectories in machine-readable form.

\subsection{Independent Samples Can Have Limited Geometric Coverage}
\label{app:ads_synthetic_independence}

For each scale $s\in\{0.05,0.10,0.25,0.50,1.00\}$, draw
$M=8$ independent samples $X_m\sim\mathcal N(0,s^2)$ per set.
We measure pairwise statistical correlation across repeated sets, with each
candidate index forming one random-variable column. This is distinct from
computing similarity between coordinate vectors within one set.
Independence gives
\begin{equation}
\operatorname{Corr}(X_i,X_j)=0\quad(i\ne j),\qquad
\operatorname{Var}\!\left[\frac{1}{M}\sum_mX_m\right]=\frac{s^2}{M}.
\label{eq:toy_independence}
\end{equation}
Thus the sample-mean effective sample size is $M$ at every positive scale.
This identity does not imply that samples occupy different spatial regions.

To measure geometric coverage separately, partition the line into cells
$B_j=[(j-\tfrac12)h,(j+\tfrac12)h)$ with $h=0.5$, and count occupied
cells in each candidate set. Also define the geometric target
$A=(0.5,\infty)$, which is an arbitrary fixed region, not a robot-success
criterion. With $b_j(s)=\Pr(X\in B_j)$ and standard normal CDF $\Phi$,
\begin{align}
\mathbb E[N_{\mathrm{cell}}]&=\sum_{j\in\mathbb Z}
\left[1-(1-b_j(s))^M\right],\\
C_M(s)&=\Pr\bigl(\exists m:X_m\in A\bigr)
=1-\Phi(0.5/s)^M.
\label{eq:toy_coverage}
\end{align}
The Gaussian cell sum is evaluated beyond ten standard deviations in either
direction. These analytic values provide a numerical check on the Monte Carlo
estimates. As $s$ approaches zero, occupancy and target coverage can decline
while statistical independence remains unchanged. This construction motivates
reporting coverage alongside any correlation-based diagnostic; it does not
assign a statistical effective sample size to semantic action diversity.

\begin{table}[htbp]
\centering
\small
\caption{Independent Gaussian sampling. Correlation is averaged over distinct candidate-index pairs across
independent sets. Coverage refers only to the predefined geometric target.}
\label{tab:toy_independence}
\begin{tabular}{c c cc cc}
\toprule
$s$ & $\widehat\rho$ & \multicolumn{2}{c}{Occupied cells}
& \multicolumn{2}{c}{Target coverage}\\
& MC & MC & Analytic & MC & Analytic\\
\midrule
0.05 & $-0.000662$ & 1.000010 & 1.000005 & 0.000000 & $6.09588\times10^{-23}$\\
0.10 & 0.000079 & 1.097250 & 1.097222 & 0.000000 & $2.29321\times10^{-6}$\\
0.25 & 0.000492 & 2.513520 & 2.512776 & 0.168490 & 0.16815\\
0.50 & $-0.000609$ & 3.644550 & 3.644815 & 0.749010 & 0.748932\\
1.00 & 0.001080 & 5.111300 & 5.116877 & 0.947410 & 0.947742\\
\bottomrule
\end{tabular}
\end{table}

\subsection{Density Weighting Does Not Determine Mode Probability}
\label{app:ads_synthetic_density}

Consider the known mixture
$f(x)=(1-w)\phi_0(x)+w\phi_1(x)$, where
$\phi_0=\mathcal N(-8,1)$ and
$\phi_1=\mathcal N(8,s_1^2)$.
We vary $w\in\{0.05,0.20,0.50\}$ and
$s_1\in\{0.01,0.10,1.00\}$.
The generating component label, rather than a fitted cluster, identifies each
sample's mode in this diagnostic. For each set of $K=8$ iid samples from $f$,
assign exact-density weights
\begin{equation}
q_k=\frac{f(X_k)}{\sum_{j=1}^{K}f(X_j)},\qquad
\widehat Q_{1,K}=\sum_{k=1}^{K}q_k\mathbf 1[Z_k=1].
\label{eq:toy_density_weights}
\end{equation}
This sets temperature to one and removes all trace-estimation error.
It tests a property of density-based reweighting itself, without assuming
that an estimated learned-flow score equals the exact density used here.

Define $I_{ab}=\int\phi_a(x)\phi_b(x)\,dx$. Gaussian products give
\begin{equation}
I_{ab}=\frac{\exp[-(\mu_a-\mu_b)^2/(2(s_a^2+s_b^2))]}
{\sqrt{2\pi(s_a^2+s_b^2)}}.
\end{equation}
The population limit of Equation~\eqref{eq:toy_density_weights} is
\begin{equation}
Q_{1,\infty}=
\frac{w(1-w)I_{01}+w^2I_{11}}
{(1-w)^2I_{00}+2w(1-w)I_{01}+w^2I_{11}}.
\label{eq:toy_population_density}
\end{equation}
For negligible overlap this reduces to
$Q_{1,\infty}\simeq(w^2/s_1)/[(1-w)^2/s_0+w^2/s_1]$.
Hence density weighting depends jointly on component probability and width;
it is not a general correction toward equal or desired mode probabilities.
Finite sets impose an additional sampling restriction:
\begin{equation}
\mathbb E[\widehat Q_{1,K}]\leq
\Pr(\exists k:Z_k=1)=1-(1-w)^K.
\label{eq:toy_finite_bound}
\end{equation}
The finite-set expectation, population limit, and probability of observing
the component are separate estimands. We report all three instead of treating
the population formula as the finite-$K$ result. No component is designated
as a more successful action in this construction.

\begin{table}[htbp]
\centering
\small
\caption{Known-mixture density weighting.
$\overline Q_{1,8}$ is a Monte Carlo mean of within-set normalized mass;
$Q_{1,\infty}$ and $P_{\mathrm{seen}}$ use the analytic formulas above.}
\label{tab:toy_density}
\begin{tabular}{cc ccc}
\toprule
$w$ & $s_1$ & $\overline Q_{1,8}$ & $Q_{1,\infty}$ & $P_{\mathrm{seen}}$\\
\midrule
0.05 & 0.01 & 0.152637 & 0.216920 & 0.336580\\
0.05 & 0.10 & 0.028886 & 0.026954 & 0.336580\\
0.05 & 1.00 & 0.003244 & 0.002762 & 0.336580\\
0.20 & 0.01 & 0.700244 & 0.862069 & 0.832228\\
0.20 & 0.10 & 0.346170 & 0.384615 & 0.832228\\
0.20 & 1.00 & 0.066806 & 0.058824 & 0.832228\\
0.50 & 0.01 & 0.981265 & 0.990099 & 0.996094\\
0.50 & 0.10 & 0.880736 & 0.909091 & 0.996094\\
0.50 & 1.00 & 0.499849 & 0.500000 & 0.996094\\
\bottomrule
\end{tabular}
\end{table}

\subsection{Actual Endpoint Updates Under a Fixed Two-Sibling Surrogate}
\label{app:ads_synthetic_endpoint}

Fix detached sibling endpoints $a_-=-1$ and $a_+=1$, target mass
$q_+\in\{0.5,0.8\}$, and scalar surrogate variance $v=0.25$.
For an optimized scalar endpoint $\mu$, the Gaussian finite-set surrogate is
\begin{equation}
p_+(\mu)=\operatorname{sigmoid}(2\mu/v),\qquad
\frac{d}{d\mu}\mathrm{KL}(q\|p(\mu))=\frac{2}{v}(p_+(\mu)-q_+).
\label{eq:toy_endpoint_gradient}
\end{equation}
For $0<q_+<1$, its unanchored minimizer is
$\mu^*=(v/2)\log(q_+/(1-q_+))$. A finite-set target can therefore be
matched by an endpoint between its two siblings; increasing a sibling's
surrogate probability is not equivalent to placing the endpoint at that
sibling or increasing the occupancy of its mode under a full generative model.

For a controlled comparison, add an explicit quadratic preference
$\frac\kappa2(\mu-1)^2$ with $\kappa\in\{0,20\}$.
This is a toy anchor, not flow matching and not a model of its full gradients.
Starting independently from $\mu_0\in\{-1,1\}$, take $400$ gradient-descent
steps of size $0.02$ on the combined scalar objective. We record the final
endpoint, its distance $d_{\min}$ to the nearest sibling, and whether it lies
within distance $0.2$ of either sibling. This defined geometric neighborhood
is not a rollout-validity label. The common anchor at $+1$ favors one endpoint; entering that neighborhood does not establish preservation of both modes. An independently computed stationary root
provides a numerical check; the companion CSV also retains every iterate.

\begin{table}[htbp]
\centering
\small
\caption{Fixed-surrogate optimization with an optional explicit quadratic
anchor. We report numerical endpoints, distances to the nearest sibling,
and geometric neighborhood indicators.}
\label{tab:toy_endpoint}
\begin{tabular}{ccc ccc}
\toprule
$q_+$ & $\kappa$ & $\mu_0$ & Final $\mu$ & $d_{\min}$ & In neighborhood\\
\midrule
0.5 & 0 & $-1$ & 0.000000000 & 1.000000000 & 0\\
0.5 & 0 & $+1$ & 0.000000000 & 1.000000000 & 0\\
0.5 & 20 & $-1$ & 0.800660032 & 0.199339968 & 1\\
0.5 & 20 & $+1$ & 0.800660032 & 0.199339968 & 1\\
0.8 & 0 & $-1$ & 0.173286795 & 0.826713205 & 0\\
0.8 & 0 & $+1$ & 0.173286795 & 0.826713205 & 0\\
0.8 & 20 & $-1$ & 0.920253802 & 0.079746198 & 1\\
0.8 & 20 & $+1$ & 0.920253802 & 0.079746198 & 1\\
\bottomrule
\end{tabular}
\end{table}

\paragraph{Interpretation boundary.}
These diagnostics isolate distinctions that a learned-policy evaluation must
address directly: independence versus spatial coverage, density versus mode
probability, and surrogate matching versus endpoint occupancy. They provide
reproducible checks of these distinctions, but cannot substitute for robot
rollouts, a comparison against compute-matched training baselines, or an
empirical study of the learned flow-score estimator.

\FloatBarrier
\section{Aggregation and Interpretation of Evaluation Metrics}
\label{app:ads_reporting}

\subsection{Task-dependent episode counts}
For task $g$ and training seed $s$, let $N_{gs}$ be the number of evaluation episodes and $Y_{gs}$ the number of successful episodes. Its task-level success percentage is $r_{gs}=100Y_{gs}/N_{gs}$. The episode-weighted aggregate for a suite or an embodiment is
\begin{equation}
R_s^{\mathrm{micro}}
=100\frac{\sum_gY_{gs}}{\sum_gN_{gs}}
=\sum_g w_{gs}r_{gs},
\qquad
w_{gs}=\frac{N_{gs}}{\sum_jN_{js}}.
\label{eq:ads_micro_average}
\end{equation}
SimplerEnv uses the official task-specific evaluation configurations and their corresponding episode counts. Since these counts differ across tasks, the reported aggregate need not equal the arithmetic mean of the displayed task success rates. The two embodiments are aggregated separately. An unweighted task mean is instead $R_s^{\mathrm{macro}}=G^{-1}\sum_g r_{gs}$; the two estimators coincide when all task counts are equal, but are otherwise different estimands. Aggregation uses the integer success and episode counts before rounding task-level percentages.

For $S_{\mathrm{seed}}$ training seeds, a mean over independently trained models is $S_{\mathrm{seed}}^{-1}\sum_s R_s^{\mathrm{micro}}$, with variation computed at the training-seed level. Pooling counts over seeds is guaranteed to be equivalent when their total episode counts are equal; otherwise it is generally a different estimator. Confidence intervals must retain the paired task instances and the grouping corresponding to the reported estimator. For benchmark perturbation groups, the same count-based definition makes the choice of group weighting explicit without changing any individual rollout outcome.

\subsection{Normalized task score and binary completion}
For real-world episode $i$, let $S_i\in[0,1]$ be the partial-completion score from Appendix~\ref{app:real_world_scoring}, and let $c_i\in\{0,1\}$ record complete execution of all four object-level subgoals. The two summary statistics are
\begin{equation}
\overline S=\frac1n\sum_{i=1}^nS_i,
\qquad
\widehat p_{\mathrm{complete}}=\frac1n\sum_{i=1}^nc_i.
\label{eq:ads_score_and_completion}
\end{equation}
Their percentages are $100\overline S$ and $100\widehat p_{\mathrm{complete}}$, respectively. A change in normalized score measures a change in partial or complete task progress; it is not numerically interchangeable with a change in binary completion probability. The sample size and uncertainty interval for each statistic are defined over rollouts, rather than over the four objects within each rollout.

\paragraph{Proposition (what a mean partial score can identify).}
Under the scoring rules in Appendix~\ref{app:real_world_scoring}, full completion has $S_i=1$. If completion fails, the four object terms are at most one each and the completion bonus is zero, so $S_i\le4/5$. Therefore
\begin{equation}
c_i\le S_i\le\frac45+\frac15c_i,
\qquad
\max\{0,5\overline S-4\}
\le\widehat p_{\mathrm{complete}}\le\overline S.
\label{eq:ads_completion_bounds}
\end{equation}
\emph{Proof.} For $c_i=1$ both sides reduce to $S_i=1$. For $c_i=0$, nonnegativity gives the lower bound and the four object terms give the upper bound. Averaging and rearranging prove the bounds on the empirical completion fraction. $\square$

These bounds generally do not identify the binary completion fraction. For example, four lifted but unsuccessfully placed objects have transfer score $4(0.8)/5=0.64$ with no complete execution. A mixture of fully completed and zero-score rollouts can have the same mean score and a different completion fraction. Episode-level completion indicators are consequently required for a binary success analysis; they cannot be reconstructed from the plotted mean score alone.

\subsection{Coverage and computational comparisons}
Coverage comparisons use an action-region definition that is fixed independently of the candidate pool being evaluated. This avoids changing the definition of a mode when the candidate budget changes. A region's geometric separation and its measured rollout validity answer different questions; neither pairwise action distance nor a clustering score certifies task success. For an oracle--verifier comparison, the candidate pool, initial-chunk execution rule, continuation policy, and outcome definition must match. Only the selection rule differs in the pointwise upper-bound interpretation in Appendix~\ref{app:ads_sampling}.

A computation-controlled training comparison holds the initialization, training data, and model-selection rule fixed, and reports both total training FLOPs and elapsed time. Matching optimizer updates alone does not match the training budget when sibling integrations and trace probes are added. An earlier-checkpoint comparison addresses retention during fine-tuning, whereas a longer-training baseline addresses additional computation; these are complementary controls. Synthetic diagnostics in Appendix~\ref{app:ads_synthetic} isolate the mathematical distinctions above and do not replace either type of learned-policy comparison.

\end{document}